\documentclass[11pt]{article}

\usepackage{booktabs}
\usepackage{graphicx}
\usepackage{amsmath}
\usepackage{xcolor}
\usepackage{natbib}
\usepackage{hyperref}
\hypersetup{hidelinks}
\usepackage{geometry}
\usepackage{caption}
\usepackage{placeins}
\usepackage{url}
\usepackage{microtype}
\graphicspath{{figures/}}

\title{Reflective Prompt Optimization of a Deep-Research Agent:\\
A Topic-Disjoint, Multi-Seed Protocol for Measuring Transferable Gains and Prompt-Level Overfitting}

\author{Elias Lumer, Sahil Sen, Vamse Kumar Subbiah, Kevin Paul}
\date{}

\begin{document}
\maketitle

\begin{abstract}
Prompt optimization is an attractive alternative to reinforcement learning for adapting
language-model agents, since it is cheap and produces human-readable artifacts. On factual tasks,
however, it carries a failure mode that scalar-reward reinforcement learning does not. An optimizer
that reads the training examples can write their answers into the prompt, so an in-distribution gain
may be memorization rather than transferable behaviour. We apply reflective prompt evolution (GEPA) to the two system prompts of a multi-agent
deep-research agent and evaluate it under a topic-disjoint split in which no topic is shared across
train, validation, and test. Held-out citation reward rises by 0.079, from 0.702 to 0.781, with a
bootstrap 95\% confidence interval of [0.051, 0.107], and the gain replicates across three seeds
(mean 0.107). Nearly half of the winning prompt is a memorized answer key, yet a controlled ablation
shows this content earns no reward even on the exact training questions it encodes. The large
validation-minus-held-out gap is therefore small-sample selection under a noisy judge, not
memorization. Varying the verifier isolates what makes the gain transfer. Decomposed per-dimension
feedback is the causal ingredient, and collapsing it to a single scalar reward makes held-out
transfer negative. The gain persists under an out-of-family judge and is largely reproduced by a
hand-written version of the discovered recipe. Credible evaluation of prompt optimizers therefore
requires held-out topics, multiple seeds, and verifier ablations. We release the candidate prompts, lineage diffs, figures, tables, and analysis.
\end{abstract}

\section{Introduction}
\label{sec:intro}

Language-model agents are increasingly adapted to downstream tasks with reinforcement learning
methods such as Group Relative Policy Optimization~\citep{grpo2024}, which typically need
thousands of rollouts and collapse a rich execution trace into a single scalar reward. Prompt
optimization is an attractive alternative because it is cheap, needs no training infrastructure,
and produces human-readable artifacts that practitioners can audit. Frameworks such as
DSPy~\citep{khattab2024dspy} have made this practical by treating a system as a program of
language-model calls and optimizing its prompts automatically, with instruction-and-demonstration
search in MIPROv2~\citep{miprov2} and reflective prompt evolution in GEPA~\citep{gepa2025}.

On factual tasks, prompt optimization has a failure mode that scalar-reward reinforcement learning
does not. An optimizer that sees the training examples can copy their specific answers into the
prompt, so the reported gain then reflects memorized lookups on seen topics rather than a
transferable change in how the agent behaves. This is a form of reward
hacking~\citep{skalse2022, amodei2016}, in which the optimizer raises the measured reward without
improving the behaviour that the reward is meant to track. It is the primary risk when a reflective
optimizer is allowed to read and rewrite prompts against a training signal. The risk is
sharpened when that signal is a rubric-style reward scored by language-model judges, which are
themselves prone to hacking in reinforcement learning with verifiable
rewards~\citep{mahmoud2025, pow3r, wang2026cherrl}, as ours is (Section~\ref{sec:reward}). A study
that reports only an in-distribution gain therefore cannot tell the reader how much of that gain
will survive on data the optimizer never saw, and closing that gap is the measurement problem we
address.

Our contribution is a generalization-first protocol that separates the two. We use a
topic-disjoint split in which no article family is shared across train, validation, and test, and
a held-out test set large enough for tight confidence intervals. Under this design the held-out
gain reports only transferable behaviour, and the difference between the validation gain and the
held-out gain lets us test whether the apparent gain is memorization. The optimizer we use as
the instrument is GEPA, the Genetic-Pareto reflective prompt evolution method of
\citet{gepa2025}. GEPA feeds a language-model reflector the full trajectory (reasoning, tool
calls, and tool outputs) together with diagnostic textual feedback, so the reflector can diagnose
why a candidate failed and propose a targeted rewrite. It keeps a Pareto frontier of candidates so
the search does not collapse to a local optimum. Because GEPA reads execution
traces, it can both discover transferable rules and copy specific facts into the prompt, which is
exactly the mixture our protocol is designed to separate. Our configuration follows recent
practice. \citet{decagon2026} report that a frontier reflection model is needed, that the
optimization sample size has a sweet spot around 20 to 100 examples, and that prompts bloat
without length regularization, and \citet{combee2026} note that enlarging the reflection batch
causes context overload. We therefore use a strong reflector and keep the train and validation
sets at 24 examples each, and we observe the bloat effect when GEPA's Pareto selection rejects its
two longest candidates.

The object we optimize is a source-attribution reward for deep-research agents~\citep{shi2025deepresearch}.
\citet{onweller2026} show that deep-research agents routinely emit citations that are present but
not verified, where the cited page exists yet fails to support the specific claim. They propose
parsing and evaluation machinery for source attribution. Our gated composite reward
(Section~\ref{sec:reward}) operationalizes those failure modes through the levers
\texttt{link\_works}, \texttt{relevant\_content}, and \texttt{fact\_check}, so optimizing against
it targets a documented attribution-quality gap. The agent's search behaviour is in turn shaped
by its harness~\citep{lumer2026rah}, and \citet{sen2026} show that harness design materially reshapes agentic search
quality, which is why prompt edits that change how the researcher searches and cites can move the
reward at all. More broadly, structured-retrieval agent systems~\citep{chronos2026} situate this
work in the line of retrieval-augmented agents~\citep{lumer2025multimodalrag}, though our task is single-shot cited reporting
rather than long-horizon conversational memory.

This paper makes the following contributions.
\begin{itemize}
  \item A generalization-first evaluation protocol, combining topic-disjoint held-out testing, a
        controlled topic-shared ablation, and multi-seed replication, that tests whether the
        validation-minus-test gap is memorization and finds that it is not. The memorized answer key
        the optimizer writes does not improve the reward, and the gap is consistent with small-sample
        selection under a noisy judge.
  \item A rigorous held-out evaluation of reflective prompt optimization on a multi-agent research
        system with a language-model-judge citation reward, with bootstrap confidence intervals
        and a paired significance test on 500 tasks, replicated across 3 optimization seeds
        (mean held-out gain 0.107, significant in every run).
  \item A controlled verifier ablation identifying decomposed per-dimension feedback as the causal
        ingredient behind transfer, since collapsing the reward to a single scalar drives held-out
        transfer negative, while finer resolution and repeated scoring do not raise it and reward
        gaming tracks the weakest verifiers rather than the strongest.
  \item A per-lever, error-analyzed account of which prompt edits drove the transferable gain and
        where the residual failures concentrate, including a prompt autopsy showing that nearly half
        of the winning prompt by word count is a memorized answer key that does not improve the
        reward, even on the exact training questions it encodes.
\end{itemize}

\section{Related work}
\label{sec:related}

\paragraph{Automatic prompt optimization.}
Automatic prompt optimization searches the natural-language input space of a fixed model rather
than its weights. Early methods treat the language model as a black-box optimizer that proposes
and scores instructions, as in APE~\citep{zhou2022ape} and OPRO~\citep{yang2024opro}, and a
parallel line borrows from evolutionary computation by mutating and recombining prompts, as in
Promptbreeder~\citep{fernando2024promptbreeder} and EvoPrompt~\citep{guo2024evoprompt}. Later work
makes the search more structured and sample-efficient, including
PromptWizard~\citep{agarwal2024promptwizard}, the study of instructions and exemplars by
\citet{wan2024}, and the joint instruction-and-demonstration search of MIPROv2~\citep{miprov2}.
GEPA~\citep{gepa2025} extends this line by reflecting on full execution traces in natural language,
and we use its reference implementation~\citep{gepacode} as the optimizer in this study. A related
text-space optimizer, SkillOpt~\citep{yang2026skillopt}, trains a persistent skill document with
bounded, validation-gated edits and an update rule that explicitly forbids hardcoding task-specific
values, a design we return to in Section~\ref{sec:limitations}.

\paragraph{Optimizing multi-module language-model programs.}
Modern agents are multi-module language-model programs, so optimizing them is more than tuning a
single instruction~\citep{bai2026maspromptbench}. The DSPy line frames these systems as declarative programs that can be compiled
and optimized~\citep{khattab2022dsp, khattab2024dspy}, and LangProbe~\citep{tan2025langprobe}
provides a benchmark for such programs. Recent methods optimize across modules, whether by combining fine-tuning
with prompt search~\citep{soylu2024}, aligning local rewards to a global objective in
Optimas~\citep{wu2025optimas}, or composing policy gradients with prompt optimization in
multi-module GRPO~\citep{ziems2025mmgrpo}. Our system is a two-module program, an orchestrator and a
researcher, and we optimize both.

\paragraph{Reflection and natural-language feedback.}
A growing body of work treats natural-language feedback, rather than a scalar, as the learning
signal. TextGrad backpropagates textual critiques through a system of language-model
calls~\citep{yuksekgonul2025textgrad}, and \citet{xu2025langfeedback} give conditions under which
such feedback provably improves a policy. Related test-time methods accumulate reusable lessons in
an external memory, as in dynamic cheatsheets~\citep{suzgun2025cheatsheet} and agent workflow
memory~\citep{wang2024awm}. These build on prompting strategies that externalize reasoning and
action, including chain-of-thought~\citep{wei2023cot} and ReAct~\citep{yao2023react}. Reflective prompt
evolution inherits the premise that a written diagnosis carries more information than a single
reward number.

\paragraph{Reinforcement learning and verifiable rewards for agents.}
The alternative to prompt optimization is reinforcement learning on the model weights. GRPO and its
relatives are effective but sample-hungry, often needing thousands of rollouts~\citep{grpo2024}, and
a line of work applies them to retrieval and search agents~\citep{xu2026traindr}, including
Search-R1~\citep{jin2025searchr1}, ReSearch~\citep{chen2025research},
R1-Searcher~\citep{song2025r1searcher}, and FrugalRAG~\citep{java2025frugalrag}, with
\citet{wu2025searchwisely} reducing wasteful searches. Because these methods need a reward,
verifiable-reward engineering has become its own topic, from generalizing verifiable instruction
following~\citep{pyatkin2025vif} to verification pipelines such as VerIF~\citep{peng2025verif}.
Rubric-style rewards are especially prone to hacking, which has motivated policy-aware rubric
design~\citep{pow3r, yang2026rubricdropout}, direct characterizations of rubric reward hacking~\citep{mahmoud2025, wang2026cherrl}, and
rubric-free distillation alternatives~\citep{rgsd}. Our reward is a language-model-judged rubric
rather than a programmatic verifier, and we optimize the prompt rather than the weights, which
avoids rollout cost but introduces the leakage risk we measure.

\paragraph{Citation faithfulness and source attribution.}
Our task is citation quality for deep-research agents, where the concern is whether a cited source
actually supports its claim. \citet{onweller2026} document that such agents routinely emit citations
that are present but unverified, and they provide parsing and evaluation machinery for source
attribution that our reward operationalizes. Faithfulness has also been pursued through training,
whether by teaching contextual faithfulness with synthetic tasks and reinforcement
learning~\citep{si2025faithfulness} or by detecting and mitigating hallucination in reasoning
models~\citep{sun2025hallucination}, and it is measured by multi-hop verification benchmarks such as
HotpotQA~\citep{yang2018hotpotqa} and HoVer~\citep{jiang2020hover}. The reliability of
language-model rubric verifiers for this exact setting, deep-research source attribution, has itself
been benchmarked against human labels~\citep{leung2026needfrontiermodelcitation}, which informs the
per-dimension judge choices in our reward. Because agent behaviour depends
heavily on the surrounding harness~\citep{sen2026, lumer2026rah} and retrieval structure~\citep{chronos2026, lumer2025multimodalrag},
prompt edits that change how an agent searches and cites can move citation quality without any
weight update. We differ from all of this work in what we measure. Prior prompt-optimization
studies report in-distribution gains, but on factual tasks such a gain conflates a transferable
behavioural change with answers the optimizer memorized from the training examples, and our
protocol separates the two, testing directly whether the reported gain survives on topics the
optimizer never saw.

\section{Method}
\label{sec:method}

\subsection{System under optimization}
\label{sec:system}

The agent is a multi-agent research system in which an orchestrator plans and delegates to a
research sub-agent~\citep{lumer2026rah} that searches a frozen corpus with a search tool, consolidates citations, and
writes a cited report. We optimize the two system prompts.

\begin{itemize}
  \item The orchestrator prompt, covering planning, delegation, synthesis, report structure, and
        citation rules.
  \item The researcher prompt, covering search strategy, evidence standards, and response format.
\end{itemize}

\subsection{Reward}
\label{sec:reward}

The reward is a gated composite in $[0,1]$, unchanged by this study, designed around the
source-attribution failure modes documented for deep-research agents~\citep{onweller2026}.

\begin{equation}
\begin{aligned}
R = \texttt{answer\_gate} \cdot \big(
&\, w_{\text{link}}\,\texttt{link\_works}
+ w_{\text{rel}}\,\texttt{relevant\_content} \\
&+ w_{\text{fact}}\,\texttt{fact\_check}
+ w_{\text{num}}\,\texttt{citation\_numbering}
\big)
\end{aligned}
\label{eq:reward}
\end{equation}

The \texttt{answer\_gate} is a correctness gate, so a wrong answer zeroes the whole reward
regardless of citation quality, and it is evaluated by a tiered check that escalates from an exact
substring match to a coverage test to a language-model judgment (Appendix~\ref{app:reward}). The
citation levers are per-citation scores averaged over a task.

\begin{itemize}
  \item \texttt{link\_works} is 1 when a cited URL exists in the corpus.
  \item \texttt{relevant\_content} measures whether the cited source is about the claim.
  \item \texttt{fact\_check} measures whether the cited source supports the specific claim and
        figures.
  \item \texttt{citation\_numbering} checks that inline references are numbered and matched to a
        source list.
\end{itemize}

The 4 weights and the verbatim judge prompts are given in Appendix~\ref{app:reward}.

\subsection{GEPA configuration}
\label{sec:gepa}

We optimize both components with the open-source GEPA implementation~\citep{gepacode}, using
Pareto candidate selection, a reflection minibatch size of 3, and a rollout budget of 240 metric
calls. GEPA is wired to the agent through a standard adapter with three functions. A rollout function runs
the agent with the candidate prompts, an evaluation function scores with the composite reward and
renders textual feedback that names each lever and surfaces the weakest citations' judge reasons,
and a record function logs traces for the reflector.

\subsection{Topic-disjoint split}
\label{sec:split}

Each task identifier encodes its source article, and questions from the same article share a
slug. We group tasks by slug and assign whole topic groups to test first, then validation, then
train, which guarantees zero topic overlap across the 3 splits
(Table~\ref{tab:splits}). Train and validation are 24 tasks over 12 topics each, and test is 500
tasks over 254 topics. The small train and validation sets sit within the 20 to 100 window
reported for GEPA~\citep{decagon2026}. The 500-task set is held-out evaluation data, not
optimization data.

\begin{table}[t]
\centering
\begin{tabular}{lrr}
\toprule
Split & Tasks & Distinct topics \\
\midrule
Train & 24 & 12 \\
Validation & 24 & 12 \\
Test & 500 & 254 \\
\midrule
Train $\cap$ Validation topics & & 0 \\
Train $\cap$ Test topics & & 0 \\
Validation $\cap$ Test topics & & 0 \\
\bottomrule
\end{tabular}
\caption{Topic-disjoint split. No article family is shared across splits, so the held-out test
set discounts any answer the optimizer memorized from train.}
\label{tab:splits}
\end{table}

\subsection{Ablation protocols}
\label{sec:ablations}

Beyond the main held-out evaluation we run two sets of controlled experiments that each vary a
single axis of the setup while holding the rest fixed. The first varies the data split and the
optimizer seed to test whether the gain is real and whether the validation-minus-held-out gap is
memorization, and the second varies the verifier to test which property of the reward signal makes
the learned rules transfer.

\paragraph{Split and seed.} A \emph{multi-seed} protocol repeats the optimization with
an identical configuration under 3 independent seeds and compares the held-out gains, which
separates the direction and significance of the effect from run-to-run variance in which candidate
the optimizer selects. A \emph{topic-shared} evaluation then measures whether the memorized answer
key actually earns reward. We hold the seed and the selected best prompt fixed, run no further
search, and score both on questions drawn from the 12 training article families. These questions are
the 24 exact training questions the prompt could have memorized, plus 96 fresh in-corpus questions
on the same families (8 per family, verbatim-quote verified and deduplicated). Because the frozen corpus is
unchanged, link resolution and search behave identically. We add no documents, only new questions
over documents the agent can already retrieve. Six families whose answers appear verbatim in the
winning prompt are flagged, so the topic-shared set decomposes into memorized and non-memorized
families (\texttt{leaky\_split\_manifest.json}).

\paragraph{Verifier configurations.} The second set of experiments holds the data, policy,
reflector, and rollout budget fixed and varies only the verifier GEPA optimizes against. The canonical
verifier uses a fine-grained reward, a single scoring pass, and per-dimension decomposed textual
feedback that names each citation lever. From it we form 3 single-knob variants: a \emph{coarse}
3-level reward resolution, \emph{3x repeated} scoring with judge averaging and a majority answer
gate, and a \emph{scalar} variant that collapses the per-dimension feedback into one aggregate
number. We run each of the 4 configurations at 2 seeds, for 8 optimization runs. We then re-score
every optimized prompt on a fixed held-out set of 120 topic-disjoint tasks with a single canonical
verifier, so that transfer is comparable across configurations regardless of the verifier each was
optimized under. Per configuration we report the validation gain under its own verifier,
the canonical held-out gain (transfer), and their difference as a gaming gap.

\subsection{Models}
\label{sec:models}

The policy agent listed in Table~\ref{tab:models} is OpenAI GPT-5.4, the deployment target, and
the reflector is OpenAI GPT-5.5 at high reasoning effort, consistent with the frontier reflector
requirement reported by \citet{decagon2026}. The reward's citation judges are
GPT-5-mini for source relevance and GPT-5.4-mini for factual support, and the answer gate uses a
language-model judge in its final tier (Appendix~\ref{app:reward}). These per-dimension judge
choices follow \citet{leung2026needfrontiermodelcitation}, who benchmark rubric language-model
verifiers against human labels for deep-research source attribution and report that a small model
suffices for source relevance while a stronger small model leads on factual support.

\begin{table}[t]
\centering
\small
\setlength{\tabcolsep}{4pt}
\begin{tabular}{lll}
\toprule
Role & Model & Note \\
\midrule
Policy (agent) & OpenAI GPT-5.4 (\texttt{gpt-5.4-2026-03-05}) & deployment target \\
Reflector (optimizer) & OpenAI GPT-5.5 (\texttt{gpt-5.5-2026-04-23}) & high reasoning effort \\
Relevance judge & OpenAI GPT-5-mini (\texttt{gpt-5-mini-2025-08-07}) & citation relevance \\
Factual-support judge & OpenAI GPT-5.4-mini (\texttt{gpt-5.4-mini-2026-03-17}) & claim support \\
Answer gate & OpenAI GPT-5-mini (\texttt{gpt-5-mini-2025-08-07}) & correctness gate (tier 3) \\
\bottomrule
\end{tabular}
\caption{Models used in the study. All identifiers are public vendor names and public API
version strings.}
\label{tab:models}
\end{table}

\section{Experiments}
\label{sec:results}

\subsection{Experimental setup}
\label{sec:expsetup}

After searching on the 24-task validation set, we evaluate the seed prompt and the best prompt on
the full 500-task held-out test set with a checkpointed and resumable runner, scoring every task
with the same gated composite reward used during optimization. We report the mean
paired reward change with a bootstrap 95\% confidence interval over tasks~\citep{efron1979bootstrap}, a Wilcoxon signed-rank
test~\citep{wilcoxon1945} and a paired $t$-test for significance, the effect size $d_z$, and the win, tie, and loss
counts. For the lever-level analysis we re-score every candidate on validation, and we re-score the
seed and best prompts on a 200-task held-out subset with the reward's per-citation reasons enabled,
which lets us decompose the reward and collect the material for the error analysis.

\subsection{Results}
\label{sec:resultssub}

\begin{figure}[t]
\centering
\includegraphics[width=0.82\textwidth]{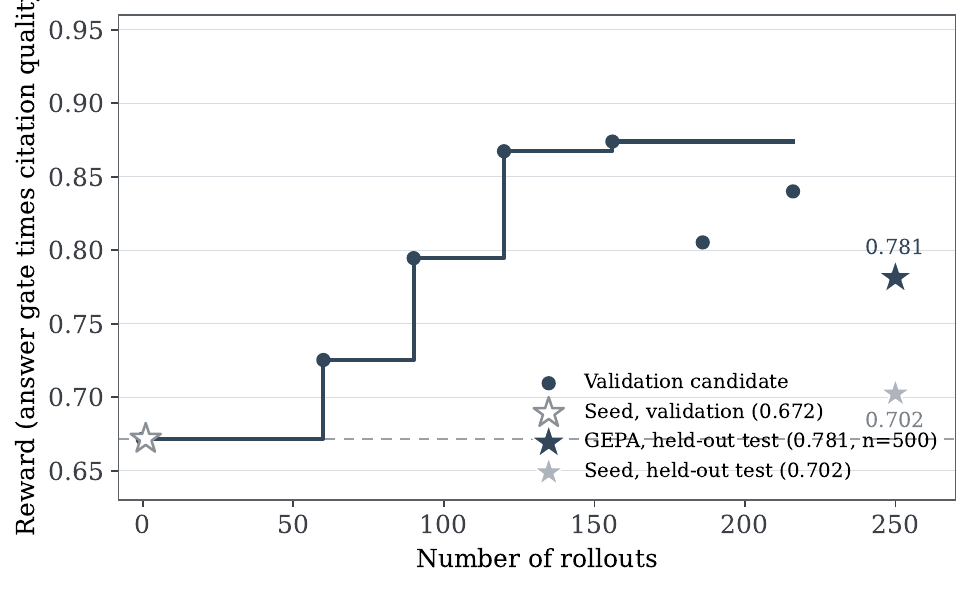}
\caption{Search trajectory. Validation reward of the best program so far against the number of
rollouts, with the seed baseline (dashed) and the held-out test stars at the right. GEPA finds
the best program by rollout 156 of 240. The held-out test gain (0.702 to 0.781) is smaller than
the validation gain, a gap we attribute to small-sample selection under a noisy judge rather than
to memorized reward (Section~\ref{sec:leakage}).}
\label{fig:rlcurve}
\end{figure}

\paragraph{Optimization trajectory.}
Over 240 rollouts GEPA produced 7 candidates (Table~\ref{tab:lineage}), and validation reward
climbs along the accepted frontier from 0.672 at the seed to 0.874 at the best candidate,
discovered at rollout 156 (Figure~\ref{fig:rlcurve}). Both prompt components contributed, with the first gain
coming from the researcher prompt and the two largest coming from the orchestrator prompt. The last
two candidates grew the prompts further and both regressed on validation, so Pareto selection kept
the shorter candidate 4. The rejection of the longer candidates 5 and 6 is in-run evidence for the
bloat effect that production users mitigate with length regularization~\citep{decagon2026},
supplied here by the Pareto objective without an explicit length penalty.

\begin{table}[t]
\centering
\small
\setlength{\tabcolsep}{5pt}
\begin{tabular}{rrrrlrrl}
\toprule
Cand. & Parent & Rollout & Val.\ reward & $\Delta$ parent & Component & Orch.\ words & Res.\ words \\
\midrule
0 & -- & 0 & 0.672 & -- & (seed) & 679 & 390 \\
1 & 0 & 60 & 0.725 & $+0.054$ & researcher & 679 & 1075 \\
2 & 1 & 90 & 0.795 & $+0.069$ & orchestrator & 1460 & 1075 \\
3 & 2 & 120 & 0.867 & $+0.073$ & researcher & 1460 & 1608 \\
\textbf{4} & 2 & 156 & \textbf{0.874} & $+0.079$ & orchestrator & 2536 & 1075 \\
5 & 4 & 186 & 0.805 & $-0.069$ & researcher & 2536 & 1503 \\
6 & 4 & 216 & 0.840 & $-0.034$ & orchestrator & 3261 & 1075 \\
\bottomrule
\end{tabular}
\caption{Candidate lineage. Each candidate mutates exactly one component of its parent.
$\Delta$ parent is the change in validation reward relative to the candidate's parent. Candidate 4
(bold) is the best program. The word counts show that candidates 5 and 6 grew longer and
regressed.}
\label{tab:lineage}
\end{table}

\begin{figure}[t]
\centering
\includegraphics[width=0.82\textwidth]{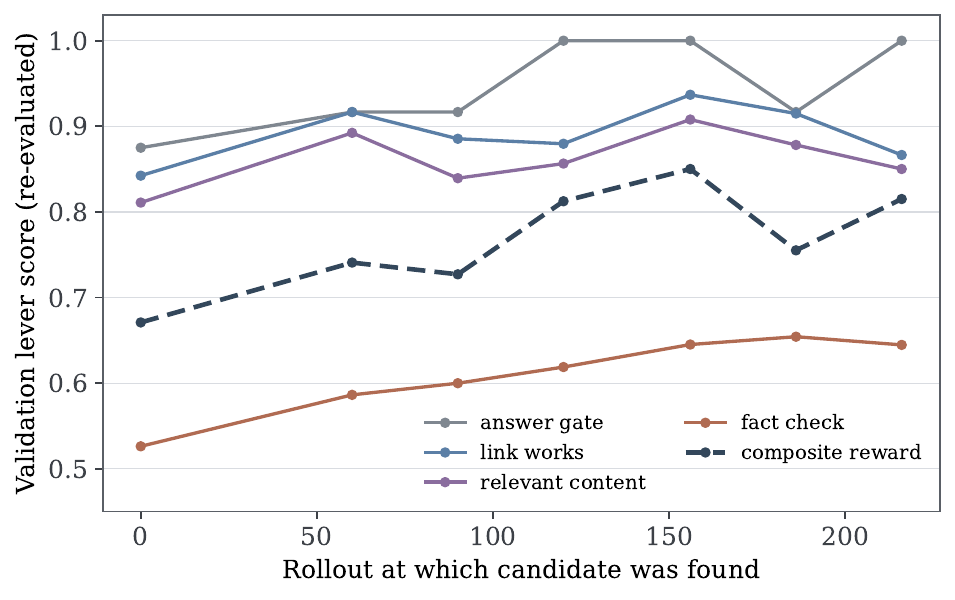}
\caption{Each reward lever as its own line over the rollout budget, from a post-hoc
re-evaluation on validation. Factual support is the lowest lever and climbs most consistently.
The composite reward (dashed) peaks at candidate 4 and dips for the over-grown candidates 5 and
6.}
\label{fig:trajectories}
\end{figure}

\begin{figure}[t]
\centering
\begin{minipage}{0.49\textwidth}
\centering
\includegraphics[width=\textwidth]{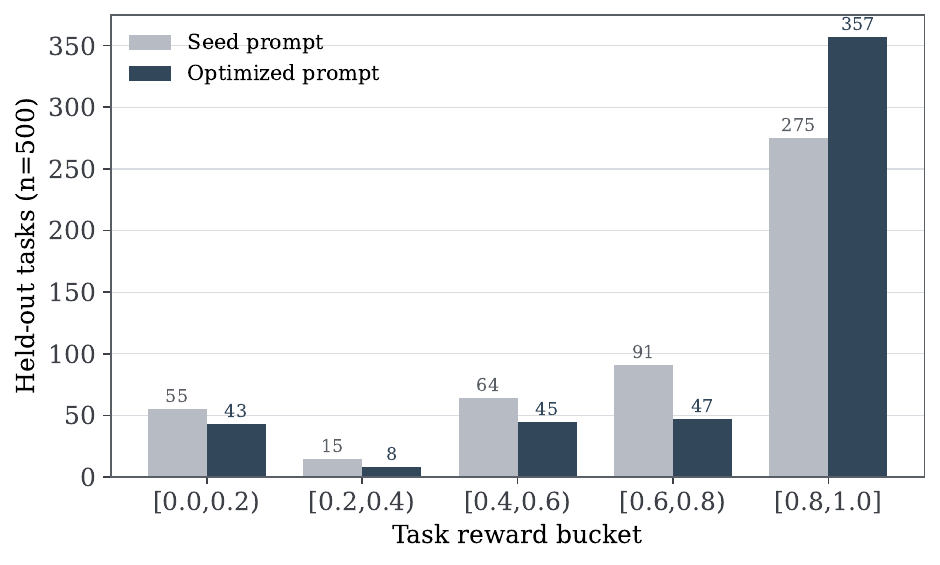}
\end{minipage}
\hfill
\begin{minipage}{0.49\textwidth}
\centering
\includegraphics[width=\textwidth]{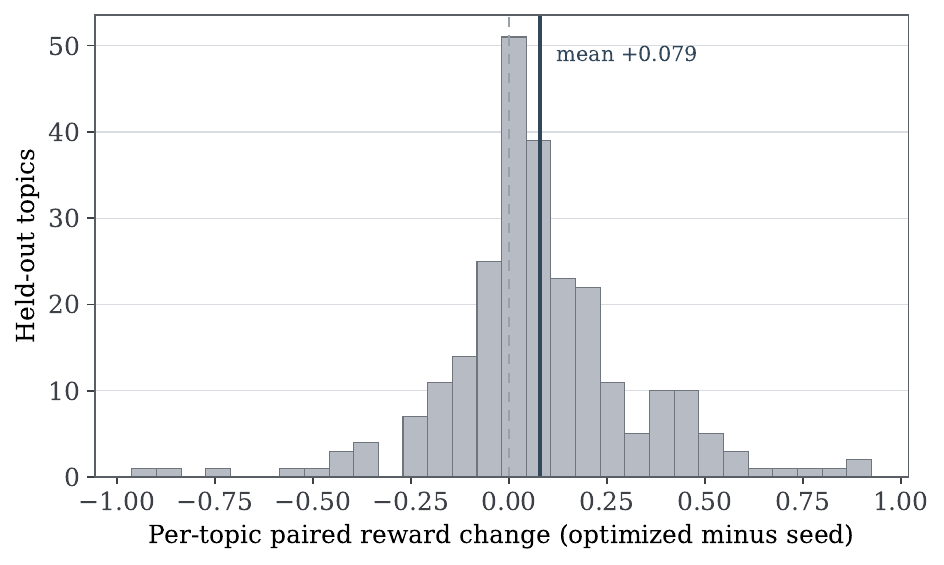}
\end{minipage}
\caption{Left: reward-bucket counts for seed and optimized across the 500 held-out tasks. The
optimized prompt shifts tasks into the top bucket. Right: distribution of the per-topic paired
reward change across the 254 held-out topics, with mean marked. 166 topics improve and 80
regress.}
\label{fig:scoredist}
\label{fig:deltahist}
\end{figure}

\begin{figure}[t]
\centering
\includegraphics[width=0.5\textwidth]{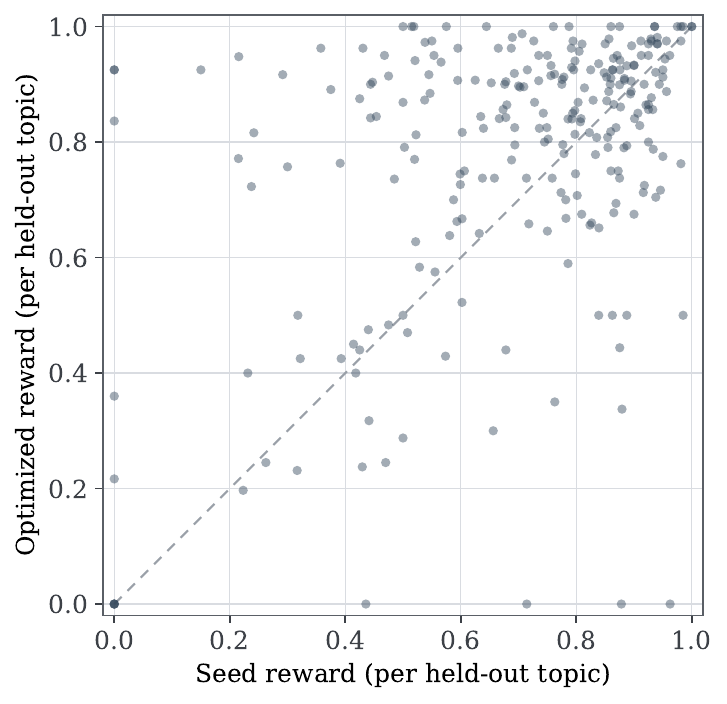}
\caption{Per-topic held-out reward, with seed on the x-axis and optimized on the y-axis and the
line of equality. The band above the line is denser, consistent with a net positive but
non-uniform gain.}
\label{fig:scatter}
\end{figure}

\paragraph{Held-out generalization.}
On the 500-task, 254-topic held-out test set the optimized prompt beats the seed by 0.079 with a
tight bootstrap confidence interval and a highly significant Wilcoxon test
(Table~\ref{tab:paired}). Because this test set shares no topic with anything the optimizer saw, it
is our headline held-out estimate. The gain is a net improvement rather than a clean sweep, with 287
tasks improving, 147 regressing, and 66 unchanged. The per-task reward distribution shifts tasks
into the top bucket (Figure~\ref{fig:scoredist}), the per-topic change is right-shifted with a
positive mean (Figure~\ref{fig:deltahist}), and the per-topic scatter is denser above the line of
equality (Figure~\ref{fig:scatter}).

\begin{table}[t]
\centering
\begin{tabular}{lr}
\toprule
Statistic & Value \\
\midrule
N (paired test tasks) & 500 \\
Distinct topics & 254 \\
Seed mean $\pm$ sd & $0.702 \pm 0.292$ \\
Optimized mean $\pm$ sd & $0.781 \pm 0.282$ \\
Mean paired delta & $+0.079$ \\
Bootstrap 95\% confidence interval & $[+0.051, +0.107]$ \\
Wilcoxon signed-rank $p$ & $6.3\times10^{-11}$ \\
Paired $t$ $p$ & $5.1\times10^{-8}$ \\
Cohen's $d_z$ & 0.247 \\
Win, tie, loss & 287, 66, 147 \\
Win rate & 57.4\% \\
Topics improved, regressed & 166, 80 \\
\bottomrule
\end{tabular}
\caption{Held-out test set, paired seed and optimized prompts (topic-disjoint, n=500).}
\label{tab:paired}
\end{table}

\begin{figure}[t]
\centering
\includegraphics[width=0.82\textwidth]{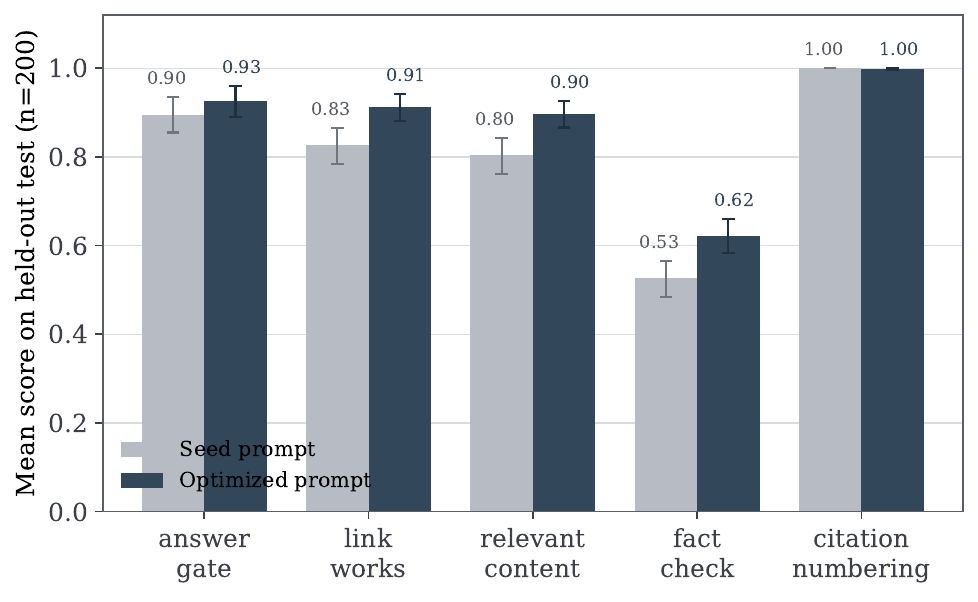}
\caption{Per-lever means on the held-out test subset (n=200) with bootstrap 95\% confidence
intervals. Link accessibility, source relevance, and factual support all improve with separated
intervals. The answer gate barely changes and citation numbering is already at ceiling.}
\label{fig:decomp}
\end{figure}

\paragraph{Reward decomposition.}
The held-out decomposition on 200 topic-disjoint tasks (Table~\ref{tab:decomp},
Figure~\ref{fig:decomp}) shows the improvement is broad across the 3 citation-quality levers,
each near 0.09 with confidence-interval-separated bars, rather than concentrated in one. Factual
support is the largest gain and, at 0.53 rising to 0.62, still the lowest lever and the binding
constraint, while the answer gate moves little and numbering is saturated. The 24-task validation
set is too noisy to decompose reliably, so we report the decomposition on the 200-task held-out set.

\begin{table}[t]
\centering
\begin{tabular}{lrrrrrr}
\toprule
Reward lever & Seed (val) & Best (val) & $\Delta$ (val) & Seed (test) & Best (test) & $\Delta$ (test) \\
\midrule
answer\_gate & 0.875 & 1.000 & $+0.125$ & 0.895 & 0.925 & $+0.030$ \\
link\_works & 0.842 & 0.937 & $+0.094$ & 0.827 & 0.913 & $+0.086$ \\
relevant\_content & 0.811 & 0.908 & $+0.097$ & 0.804 & 0.898 & $+0.094$ \\
\textbf{fact\_check} & 0.526 & 0.645 & $+0.119$ & \textbf{0.527} & \textbf{0.622} & \textbf{+0.095} \\
citation\_numbering & 1.000 & 1.000 & $+0.000$ & 1.000 & 0.998 & $-0.002$ \\
\bottomrule
\end{tabular}
\caption{Reward decomposition per lever, seed and best prompts. Validation columns are n=24 and
held-out test columns are n=200. Factual support is the largest held-out gain and still the
lowest lever.}
\label{tab:decomp}
\end{table}

\subsection{Robustness across seeds}
\label{sec:seeds}

Repeating the optimization under 3 independent seeds shows the held-out gain is a property of the
method, not of one lucky run (Table~\ref{tab:seeds}). Every seed produces a positive held-out gain
whose bootstrap confidence interval excludes zero, with a mean of 0.107 and a range of 0.079 to
0.130. What is not stable is the validation-minus-held-out gap, which ranges from 0.034 to 0.160, a
factor of about 5. A quantity that measured the memorized portion of the same selected prompt should
not swing this much across seeds, so the spread is direct evidence that the gap reflects
optimizer-selection and judge variance rather than a stable memorization signal.

\begin{table}[t]
\centering
\small
\setlength{\tabcolsep}{5pt}
\begin{tabular}{lrrrlrl}
\toprule
Seed & Val.\ $\Delta$ & Test $\Delta$ & Bootstrap 95\% CI & Wilcoxon $p$ & Win/tie/loss & Val$-$test gap \\
\midrule
0 & $+0.202$ & $+0.079$ & $[+0.051, +0.107]$ & $6.3\times10^{-11}$ & 287/66/147 & 0.123 \\
1 & $+0.290$ & $+0.130$ & $[+0.104, +0.156]$ & $3.0\times10^{-29}$ & 348/39/113 & 0.160 \\
2 & $+0.147$ & $+0.112$ & $[+0.086, +0.140]$ & $3.4\times10^{-17}$ & 309/46/145 & 0.034 \\
\midrule
mean & -- & $+0.107$ & -- & -- & -- & -- \\
\bottomrule
\end{tabular}
\caption{Multi-seed robustness on the topic-disjoint held-out set (n=500 per seed). The held-out
gain is positive and significant in every run (mean 0.107, sd 0.021). The validation-minus-held-out
gap is not stable, ranging from 0.034 to 0.160.}
\label{tab:seeds}
\end{table}

\subsection{Does the memorized answer key earn reward?}
\label{sec:leaky}

The topic-shared evaluation isolates the effect of the memorized answer key by scoring the fixed
seed and best prompts on questions from the training families, with no further search
(Table~\ref{tab:leaky}). On the 24 exact training questions, which is the best possible case for a
memorized lookup, the best prompt beats the seed by only 0.090. That gain is statistically
indistinguishable from the 0.079 held-out gain and nowhere near the 0.202 validation gain, and the
answer gate stays flat at 0.92 for both prompts, so the memorized key does not even raise answer
recall. If memorization drove the validation gain, this exact-leak number would approach 0.202, but
it does not. On fresh same-topic questions the memorized families gain less (0.071) than the
non-memorized families (0.154), so the answer-key blocks behave as prompt bloat that slightly
depresses performance rather than as a useful lookup table. Several memorized entries do not even
match the dataset's gold answers, which reinforces that the key is not a reliable table of answers.
A topic-shared split does inflate the apparent gain modestly (0.112 versus the 0.079 disjoint gain),
but the inflation is the transfer of general citation discipline to same-topic questions, not
memorized reward, and it remains far below the validation gain.

\begin{table}[t]
\centering
\small
\setlength{\tabcolsep}{5pt}
\begin{tabular}{lrrrrr}
\toprule
Subset & n & Seed & Best & $\Delta$ & Win/tie/loss \\
\midrule
Exact-leak (all 24 train questions) & 24 & 0.695 & 0.784 & $+0.090$ & 12/6/6 \\
\quad memorized families & 12 & 0.606 & 0.695 & $+0.089$ & 6/2/4 \\
\quad other train families & 12 & 0.783 & 0.874 & $+0.090$ & 6/4/2 \\
Topic-shared fresh (all 96) & 96 & 0.593 & 0.705 & $+0.112$ & 53/15/28 \\
\quad memorized families & 48 & 0.559 & 0.630 & $+0.071$ & 25/10/13 \\
\quad non-memorized families & 48 & 0.627 & 0.781 & $+0.154$ & 28/5/15 \\
\bottomrule
\end{tabular}
\caption{Topic-shared evaluation of the fixed seed and best prompts, no further search. On the exact
training questions the memorized answer key adds only 0.090, indistinguishable from the disjoint
held-out gain and far below the 0.202 validation gain, and it helps memorized families less than
non-memorized ones. Each cell is a single rollout per task, so small cells (n=12) carry variance;
the robust reading is the pattern.}
\label{tab:leaky}
\end{table}

\subsection{Which verifier property drives transfer?}
\label{sec:verif}

\begin{figure}[t]
\centering
\includegraphics[width=0.7\textwidth]{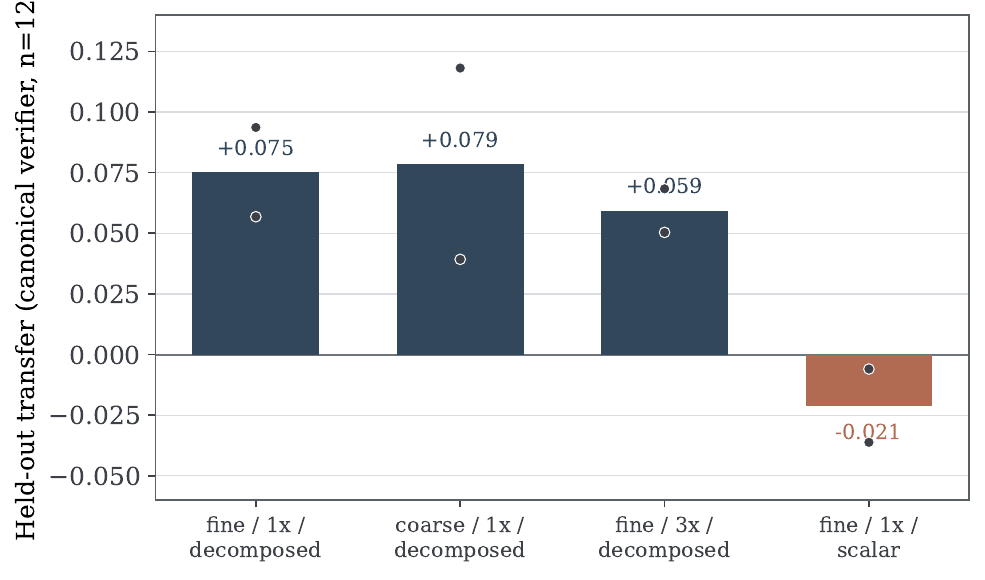}
\caption{Held-out transfer (canonical verifier, n=120) for each verifier configuration, mean of 2
seeds with the individual seeds overlaid. Decomposed per-dimension feedback transfers, while
collapsing it to a scalar reward drives transfer negative.}
\label{fig:veriftransfer}
\end{figure}

This ablation varies the verifier while holding the data, policy, reflector, and budget fixed, and
re-scores every optimized prompt on one canonical held-out verifier so the configurations are
comparable (Table~\ref{tab:verif}, Figure~\ref{fig:veriftransfer}). It varies three properties of
the reward signal that supervises the search, its resolution, whether scoring is repeated, and
whether the feedback is decomposed per dimension, and asks which of them determines what transfers
and where reward gaming appears. The canonical held-out baseline is 0.682 over 120 topic-disjoint
tasks, and each configuration is 2 seeds, so we read the cell means as directional rather than
precise (Section~\ref{sec:limitations}).

\paragraph{Decomposed feedback is what makes the gain transfer.} Replacing the per-dimension textual
feedback with a single scalar reward drives held-out transfer negative, from $+0.075$ for the
decomposed reference to $-0.021$ for the scalar variant. Per seed the scalar reward either found no
accepted improvement at all or raised validation while regressing on held-out, so a reflector given
only a scalar has nothing to diagnose and cannot discover transferable citation-discipline rules.
The levers that carry the transferable gain in the main results, \texttt{link\_works},
\texttt{relevant\_content}, and \texttt{fact\_check}, are exactly the per-dimension signals that the
scalar variant discards.

\paragraph{Finer or repeated scoring buys stability, not transfer.} Along the reliability ordering
of coarse, fine, and fine with 3x repetition, mean transfer is flat to declining at 0.079, 0.075,
and 0.059, so more granular or more repeated judging did not improve generalization at a fixed
budget, and repetition cost 3x the judge calls. Its one benefit was stability, since repeated
scoring gave the tightest seed-to-seed spread at 0.018, against 0.037 for the reference and 0.079
for the coarse verifier.

\begin{figure}[t]
\centering
\includegraphics[width=0.72\textwidth]{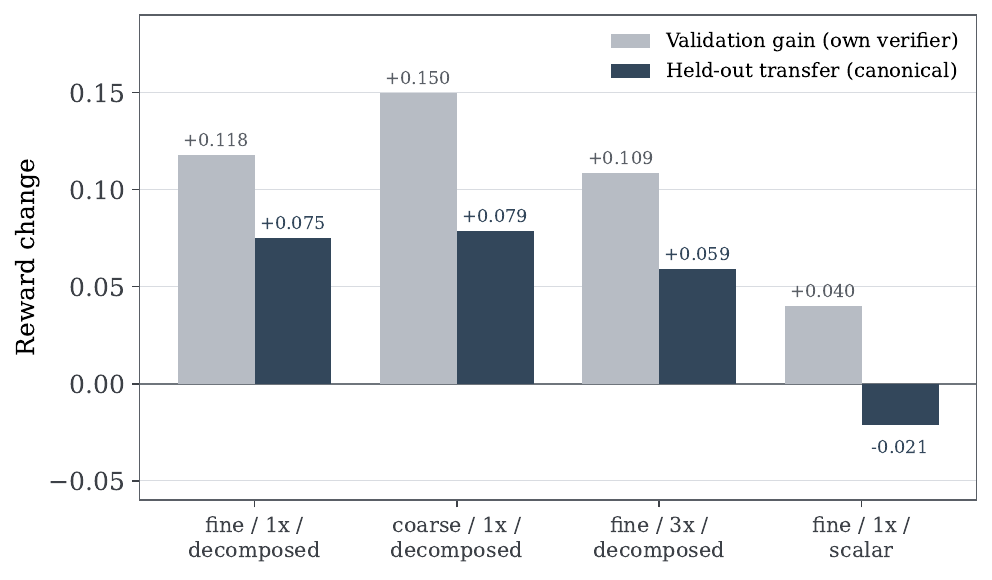}
\caption{Validation gain under each configuration's own verifier against canonical held-out
transfer. The gap between the two, the gaming gap, is largest for the coarse and scalar verifiers,
and held-out re-scoring exposes it.}
\label{fig:verifgap}
\end{figure}

\paragraph{Reward gaming tracks weak verifiers, and held-out testing exposes it.} The gaming gap,
the difference between the validation gain and the held-out transfer, is largest for the
low-information verifiers rather than the strongest one, at 0.071 for coarse and 0.061 for scalar on
top of its negative transfer, against 0.043 for the fine decomposed reference
(Figure~\ref{fig:verifgap}). Stronger verification therefore did not induce more gaming in this
setting, low-information verification did, and the held-out re-scoring is what reveals it, most
starkly for the scalar cell where a positive validation delta hides a held-out regression.

\begin{table}[t]
\centering
\small
\setlength{\tabcolsep}{6pt}
\begin{tabular}{lrrrr}
\toprule
Verifier configuration & Val.\ $\Delta$ & Transfer $\Delta$ & Gaming gap & Seed spread \\
\midrule
fine / 1x / decomposed (reference) & $+0.118$ & $+0.075$ & 0.043 & 0.037 \\
coarse / 1x / decomposed & $+0.150$ & $+0.079$ & 0.071 & 0.079 \\
fine / 3x / decomposed & $+0.109$ & $+0.059$ & 0.049 & 0.018 \\
fine / 1x / scalar & $+0.040$ & $-0.021$ & 0.061 & 0.030 \\
\bottomrule
\end{tabular}
\caption{Verifier-scaling ablation, means over 2 seeds, all re-scored on one canonical held-out
verifier (n=120, baseline 0.682). Transfer is the canonical held-out gain and the gaming gap is the
validation gain minus that transfer. Only removing decomposed feedback (scalar) changes the sign of
transfer. Finer or repeated scoring does not raise it, and the gaming gap is largest for the weakest
verifiers.}
\label{tab:verif}
\end{table}

\subsection{A hand-written recipe baseline}
\label{sec:recipe}

The transferable rules the optimizer found are describable, which raises the question of whether a
practitioner could write them by hand and skip the search, so we test that directly. We construct a
\emph{recipe} prompt equal to the best prompt with all memorized answer-key content removed, dropping
the \texttt{Known Benchmark Facts} and \texttt{Known Prior Research Hints} blocks and the two
answer-bearing citation examples, and keeping only the describable rules of answer-first structure,
one claim per citation, verified page content, corpus-only URLs, and sequential numbering. We score
this recipe against the seed and the best prompt on 100 held-out tasks under the same policy, reward,
and corpus (Table~\ref{tab:recipe}).

The recipe alone recovers 0.047 of the 0.083 total held-out gain on these tasks, about 57\%, so a
practitioner who writes down the rules captures most of the improvement. The full optimized prompt
adds a further 0.036, but this increment is not individually significant at 100 tasks (95\%
confidence interval $[-0.021, +0.096]$), consistent with the memorized answer key being inert
(Section~\ref{sec:leaky}). The total seed-to-best gain of 0.083 is significant and matches the
headline 0.079 on the full 500-task set. The improvement is therefore largely a transferable,
describable recipe, and reflective search adds only a small residual that we cannot separate from
noise at this sample size.

\begin{table}[t]
\centering
\small
\begin{tabular}{lrrr}
\toprule
Contrast & $\Delta$ held-out & Bootstrap 95\% CI & Win/tie/loss \\
\midrule
Recipe $-$ seed & $+0.047$ & $[-0.011, +0.106]$ & 51/14/35 \\
Best $-$ recipe & $+0.036$ & $[-0.021, +0.096]$ & 43/18/39 \\
Best $-$ seed & $+0.083$ & $[+0.019, +0.148]$ & 56/12/32 \\
\bottomrule
\end{tabular}
\caption{Hand-written recipe baseline on 100 held-out tasks, where the recipe is the best prompt with
all memorized answer-key content removed. Mean held-out reward is 0.684 (seed), 0.731 (recipe), and
0.767 (best). The describable recipe recovers about 57\% of the total gain, and the residual from the
full search is not individually significant at this sample size.}
\label{tab:recipe}
\end{table}

\subsection{Cost and latency}
\label{sec:cost}

Because prompt optimization changes only the prompt, the optimized agent runs at inference with no
added machinery, and we confirm it carries no latency penalty. On 6 held-out queries per prompt at
concurrency one, the optimized prompt averaged 138.8 seconds per query against 159.9 for the seed,
and output length was unchanged at about 3200 tokens. The gain does cost input tokens, since mean
total policy tokens rose about 32\%, from 146.0k to 193.1k per query, almost entirely on the input
side (an added 47k), because the optimized prompt retrieves and grounds more context per query.
GEPA's search is a one-time cost of 240 policy rollouts, on the order of 40M policy tokens, after
which every query runs at the per-query cost above with no recurring optimizer overhead. These are
small-sample point estimates over 6 queries per prompt, reported for scale rather than as a powered
comparison.

\subsection{Judge validation and self-preference}
\label{sec:judgeval}

The reward is scored by language-model judges~\citep{zheng2023judge}, so we check that they agree with a human and are not
inflated by a shared provider family~\citep{panickssery2024selfpref, pombal2026selfpref}. One author labeled 48 held-out (attribution, citation) pairs
blind to the judge decision, and we re-scored the same pairs with an out-of-family judge, Anthropic
Claude Opus 4.8 (Table~\ref{tab:judgeval}). The in-family factual-support judge, which scores the
binding reward lever, matched the human on 80.0\% of pairs (Cohen's $\kappa = 0.60$; \citealp{cohen1960kappa}), and its
disagreements are mildly lenient, accepting topically-close sources on 7 of its 9 mismatches. The
relevance judge matched on 77.8\% ($\kappa = 0.23$), and all 10 of its disagreements are
one-directional under-crediting, where it withholds relevance because the source does not also
support the specific fact, so it never inflates the score. The out-of-family judge agrees with the
human as well as or better than the in-family judge, on 91.1\% of relevance and 80.0\% of
factual-support decisions, and where the two families differ it is the in-family judge that is the
more lenient. Re-scoring the composite with the out-of-family judge on 12 held-out tasks, the
seed-to-best gain persists and is if anything larger, 0.129 against 0.069 in-family on the same
tasks, so the improvement is not an artifact of the judges' provider family. These are spot-checks
with 48 pairs and a single annotator, and they complement the larger judge-versus-human benchmark of
\citet{leung2026needfrontiermodelcitation}.

\begin{table}[t]
\centering
\small
\begin{tabular}{llrrr}
\toprule
Comparison & Dimension & n & Agreement & Cohen's $\kappa$ \\
\midrule
Human vs.\ in-family judge & source relevance & 45 & 77.8\% & 0.23 \\
Human vs.\ in-family judge & factual support & 45 & 80.0\% & 0.60 \\
Human vs.\ out-of-family judge & source relevance & 45 & 91.1\% & 0.46 \\
Human vs.\ out-of-family judge & factual support & 45 & 80.0\% & 0.61 \\
\bottomrule
\end{tabular}
\caption{Judge validation on held-out (attribution, citation) pairs. The in-family judges agree with
a human annotator, most strongly on factual support, the binding lever. An out-of-family judge
(Anthropic Claude Opus 4.8) agrees with the human as well as or better than the in-family judge, and
the seed-to-best gain persists under it ($+0.129$ against $+0.069$ in-family on 12 tasks).}
\label{tab:judgeval}
\end{table}

\section{Discussion}
\label{sec:leakage}

The winning lineage rewrote the researcher prompt once and the orchestrator twice. Across these
edits the reflector added topic-independent rules, each targeting a different citation lever.
It moved the answer to the front of the report, split multi-claim sentences so that one claim
carries one citation, required the agent to verify that the cited page's main content supports the
specific claim, and restricted citations to corpus URLs. These edits are what raise the 3
citation levers on held-out data.

The verifier ablation (Section~\ref{sec:verif}) explains why the reflector could find these
transferable rules at all. Each of them answers a specific diagnostic the decomposed reward
surfaces, that a link failed to resolve, that a cited page was off-topic, or that a page did not
support the claim, and the reflector rewrites the prompt to close the named failure. When the same
reward is collapsed to a single scalar the reflector loses that diagnostic channel, and transfer
turns negative, so the mechanism behind the gain is the per-dimension feedback rather than the
optimizer's search machinery on its own. This also reframes the reward signal as the design
variable that matters most, since a practitioner who reports only a scalar would both learn less and
be more exposed to the gaming that the held-out protocol exists to catch.

The reflector also made a second kind of edit. It wrote \texttt{Known Benchmark
Facts} and \texttt{Known Prior Research Hints} sections that memorized specific training answers,
for example a language count for one model, a product name for one company, and a policy date for
one central bank. This is a textbook case of prompt-level overfitting~\citep{fu2026textreg}, in which the optimizer copies
specific training values into the prompt instead of learning a rule that generalizes, an attempted
reward hack against the training signal~\citep{skalse2022, mahmoud2025}. This memorization is not confined to the candidates that
Pareto selection rejected. It appears in the first accepted edit and grows to dominate the winning
program, in which the memorized answer key is 52\% of the orchestrator prompt and 39\% of the
researcher prompt by word count. Nearly half of the deployed program is therefore a table of
verbatim training questions paired with their exact answers and source URLs
(Table~\ref{tab:memshare}). The
natural reading is that this key is what inflates the validation score, but the topic-shared
ablation (Section~\ref{sec:leaky}) shows it is not. The memorized key adds only 0.090 even on the
exact training questions it encodes, which is indistinguishable from the transferable held-out gain
and far below the 0.202 validation gain, and it helps the memorized families less than the
non-memorized ones. The memorized content is therefore attempted reward hacking that does not pay
off, raising the prompt's length far more than its reward, and the validation-minus-held-out gap is
better explained by small-sample selection under a noisy judge, which we take up next.

\begin{table}[t]
\centering
\begin{tabular}{lrr}
\toprule
Candidate & Orchestrator & Researcher \\
\midrule
0 (seed) & 0\% & 0\% \\
1 & 0\% & 39\% \\
2 & 40\% & 39\% \\
3 & 40\% & 40\% \\
\textbf{4 (best)} & \textbf{52\%} & \textbf{39\%} \\
5 & 52\% & 72\% \\
6 & 54\% & 39\% \\
\bottomrule
\end{tabular}
\caption{Share of each candidate prompt, by word count, occupied by memorized
\texttt{Known Benchmark Facts} and \texttt{Known Prior Research Hints} answer-key sections that
pair verbatim training questions with their answers. Memorization enters at the first accepted
edit and occupies nearly half of the selected best program (candidate 4). The topic-shared ablation
(Section~\ref{sec:leaky}) shows this content does not improve the reward, even on the exact training
questions it encodes.}
\label{tab:memshare}
\end{table}

Prompt bloat is a subtler symptom of the same pressure. Past candidate 4 the reflector kept adding
emphatic and lengthy instruction, more capitalized \texttt{CRITICAL} directives and more
memorized-fact dumps, growing the orchestrator prompt to 3261 words in candidate 6, yet both
over-grown candidates regressed on validation (Table~\ref{tab:lineage}). The extra instruction
inflated length without improving behaviour, and GEPA's Pareto selection rejected it, a useful
partial guard against reward hacking by bloat, but only a validation-time one. It is the
topic-disjoint held-out test, not the Pareto objective, that ultimately separates the transferable
recipe from the memorized and over-instructed content.

\subsection{Why the topic-disjoint protocol is necessary}
\label{sec:necessity}

The protocol has three parts, and each catches a failure the others miss. The topic-disjoint split
is what makes a held-out score mean transfer, because no article family is shared across train,
validation, and test, so a memorized training answer cannot inflate the held-out number. Held-out
evaluation then converts the apparent validation gain into a transfer estimate, and the gap between
the two measures the part that does not transfer. A practitioner who selected and reported on
validation alone would have described a gain of 0.202 when the transferable gain is 0.079, an
overstatement of about 2.6 times.

The gap is not, however, a stable measurement of memorized reward, and two results establish this.
Across the candidates the memorized answer-key fraction and the validation reward rise together
(Pearson $r = 0.83$), from 0\% at the seed to 48\% at the selected best
(Table~\ref{tab:memshare}, Table~\ref{tab:lineage}), which invites the reading that the optimizer is
buying validation reward with memorization. That reading mistakes correlation for causation, because memorization
and the transferable citation-discipline edits were added along the same accepted lineage, so the
two rise together. The topic-shared ablation (Section~\ref{sec:leaky}) nonetheless shows the
memorized content earns no reward even on the questions it encodes. What remains true is that a validation-only
workflow surfaces the most-memorized program as its best, so the practitioner not only mis-sizes the
gain but also ships the most overfitted prompt, even though that overfitting is chasing judge noise
rather than memorized reward.

The prompt autopsy supplies qualitative ground truth for the overfitting itself. Reading the winning
prompt confirms that its \texttt{Known Benchmark Facts} sections pair specific training questions
with their exact answers, for example a fixed language count for one model and a fixed policy date
for one central bank. The overfitting is therefore literal memorization rather than benign
paraphrase. The
autopsy establishes that the memorized content is real and large, and the ablation establishes that
it is non-functional. Together they separate a genuine prompt-hygiene problem from the size of the
transfer gap.

The topic-shared counterfactual is no longer only a prediction. The ablation in
Section~\ref{sec:leaky} constructs same-family questions and scores the fixed prompts on them, and
it does not collapse the gap toward the validation number. The apparent gain rises only to 0.112,
well short of the 0.202 validation gain, and the memorized families gain less than the
non-memorized ones. Because the gap is dominated by small-sample selection under a noisy judge, we
expect it to be largest not where memorization is heaviest but where the validation set is smallest
and the judge noisiest. We separately expect memorization pressure to be largest in narrow,
stationary domains where the same entities recur (Section~\ref{sec:limitations}).

\subsection{Error analysis}
\label{sec:errors}

After optimization, factual support is still the lowest lever at 0.62 on the held-out subset, and
59 of 200 tasks score below 0.5 on it. Table~\ref{tab:failthemes} categorizes the weak citations
by the judge's stated reason. The dominant theme is that the cited source does not support the
specific claim, consistent with factual support being the binding lever. The corpus-only-URL rule
largely worked, since not-in-corpus URLs are nearly gone.

\begin{table}[t]
\centering
\begin{tabular}{lr}
\toprule
Failure theme & Weak-citation count \\
\midrule
Source does not support the specific claim & 173 \\
Topically related but not the exact fact & 165 \\
Composite or multi-claim citation & 36 \\
Wrong number, figure, or date & 6 \\
URL inaccessible or not in corpus & 4 \\
\bottomrule
\end{tabular}
\caption{Fact-check failure themes for the optimized prompt on the held-out subset, from the
per-citation judge reasons.}
\label{tab:failthemes}
\end{table}

The regressions are mostly answer-gate flips, where the optimized prompt is more assertive because
it puts the answer first, which usually helps but sometimes commits to a wrong specific answer
where the seed hedged. On the paired 200-task subset the gate gained on 12 tasks and lost on 6. The gate losses
are the largest single-task regressions, where a task can drop from near 0.9 to 0, but they are
outnumbered by the gate gains, and the mean factual-support change on the subset is positive at
0.095.

\section{Limitations and Future Work}
\label{sec:limitations}

\paragraph{The reward is a language-model judge, validated only at small scale here.}
Every number in this paper is measured by the composite reward, which is itself scored by
language-model judges. A spot-check of 48 held-out pairs (Section~\ref{sec:judgeval}) shows the
judges agree with a human, most strongly on the binding factual-support lever ($\kappa = 0.60$), and
the per-dimension judge models were validated more broadly against human labels in a companion
benchmark~\citep{leung2026needfrontiermodelcitation}. Our own check is single-annotator and small, so
a larger multi-annotator study on the optimized outputs would strengthen it, but the reported gain is
a gain against judges of demonstrated human agreement rather than an unaudited proxy.

\paragraph{Self-preference and same-family bias.} The policy, the reflector, and the citation
judges are all from the same provider family, which risks the reflector learning to please the
judge's scoring idiosyncrasies and same-family agreement inflating the measured reward~\citep{panickssery2024selfpref, pombal2026selfpref}. We check this
directly in Section~\ref{sec:judgeval}, where an out-of-family judge agrees with the human as well as
or better than the in-family judge and the seed-to-best gain persists under it, so the improvement is
not an artifact of the shared family. The check uses one alternative model on a small sample, so a
larger multi-provider judge panel would strengthen it, though the current evidence points against a
self-preference confound.

\paragraph{Constraining the optimizer against memorization.} Our protocol detects prompt-level
overfitting but does not prevent it. The reflector is free to write anything into the prompt,
including the memorized answer key of Section~\ref{sec:leakage}, and the topic-disjoint split only
discounts that content at evaluation rather than stopping it at the source. Here the memorized key
turned out not to improve the reward, so constraining it is a matter of prompt hygiene and
auditability rather than recovering lost accuracy, but in a domain where such content did transfer
the same edit would be an active hazard. A complementary direction is to
constrain the update step itself~\citep{fu2026textreg}. SkillOpt~\citep{yang2026skillopt} enforces generalizability
inside the optimizer prompt, instructing the editor that edits must be generalizable and must not
hardcode task-specific values, filtering example-specific proposals before they are applied, and
accepting only bounded edits that pass a held-out validation gate. A generalization-constrained,
validation-gated update rule of this kind could suppress memorized-answer-key edits during
optimization instead of leaving them to be caught after the fact, and pairing such a rule with our
protocol is a natural next step.

\paragraph{Small-sample validation and the winner's curse.} The optimizer selects its best
candidate on 24 validation tasks scored by a stochastic judge, so the reported validation reward of
the winner is subject to winner's-curse inflation. Selecting the maximum over several noisy
candidates favours the one that caught the most upward noise, a regressional-Goodhart
effect~\citep{manheim2018goodhart, gao2023overopt}. This is the most likely source of the validation-minus-held-out gap,
consistent with the gap's instability across seeds (0.034 to 0.160, Section~\ref{sec:seeds}) and
with the ablation showing the memorized content earns no reward (Section~\ref{sec:leaky}). A larger
validation set, an explicit selection penalty, or repeated scoring to average out judge noise would
shrink the gap, but we did not run these and leave them for future work.

\paragraph{Reward engineering as a complementary guard against hacking.} A complementary intervention point is the reward function itself. Verbatim answer injection is detectable at scoring time: a prompt-content novelty penalty, subtracting a term proportional to token overlap between the candidate prompt and training questions, would directly penalize this strategy in-run rather than catching it post-hoc via the held-out split. In addition, a judge could be prompted to identify leakage. More broadly, reward ensembles have shown promise in improving robustness against reward hacking~\citep{coste2024ensembles}, and the same principle applies to prompt optimization: scoring candidates against multiple independent reward heads makes it harder to inflate any single proxy. Designing more robust and tamper-resistant reward functions and detecting misalignment via behavioral or distributional anomaly detection~\citep{yang2026hackprobe} are both active directions for mitigating reward hacking, and adapting these techniques to the prompt-optimization setting, where the policy becomes the prompt instead of model weights, is a natural extension of the present work \citep{reward_survey}.

\paragraph{Three seeds, not many.} The held-out gain replicates across 3 independent seeds
(Section~\ref{sec:seeds}), so its direction and significance are robust rather than a single lucky
run. What varies is the magnitude of the gain (0.079 to 0.130) and which candidate the optimizer
selects, and 3 seeds is too few to characterize that spread precisely. We therefore do not treat the
validation-minus-held-out gap as a stable estimate of anything. It is a diagnostic that flags
overfitting, not a calibrated measurement.

\paragraph{The verifier ablation is directional.} The verifier ablation
(Section~\ref{sec:verif}) runs each of the 4 configurations at only 2 seeds, and the seed-to-seed
spreads are wide. We therefore read the cell means as directional rather than as tight estimates and
do not lean on the exact ordering among the decomposed cells. The robust readings are the two that survive
the spread. Scalar feedback collapses transfer, and there is no reliability-to-transfer trend. All
cells are re-scored by a single canonical verifier from one provider family on a 120-task held-out
subset, so the transfer numbers inherit that judge's biases, and confirming the mechanism with an
out-of-family verifier and more seeds is the natural follow-up.

\paragraph{Reward under-specification and corpus binding.} The reward weights, the tiered
answer-gate definition, and the judge prompts are released with the artifact but should also be
stated in the paper. The search tool queries a frozen corpus, so the link-accessibility and
URL-discipline conclusions are relative to that corpus.

\paragraph{A single agent, task, and domain.} We study one multi-agent deep-research agent on one
citation task drawn from a single news corpus, so the size of the leakage and the exact
transferable recipe are specific to this setting. Whether reflective optimization leaks as heavily
on other tasks, agents, and domains is open, and we expect the memorization pressure to be larger,
not smaller, in narrow domains. A domain with a small and recurring entity space, a fixed document
template, or a reward that keys on a few specific facts gives the optimizer more opportunity to bake
answers into the prompt and more reward for doing so. Replicating the protocol on such a domain,
with splits that are disjoint by entity or document source rather than by news topic, is the natural
test of external validity.

\paragraph{Held-out monitoring for deployed prompt optimization.} In this setting the gap was
dominated by selection noise, but the same instrumentation is what would catch the more dangerous
case in which memorized content does transfer. Recomputing the held-out gain on a schedule against a
refreshed, entity-disjoint set, and pairing it with a periodic prompt autopsy, gives two
complementary signals for a deployed prompt-optimized system. The held-out score tracks whether the
gain still generalizes, and the autopsy flags when the optimizer has started baking specific entities
into the prompt. In a stationary domain a baked-in answer keeps scoring well until the entity it
names changes, at which point the held-out score falls while the in-distribution score does not.
Tracking these signals over time, rather than treating any single validation-minus-held-out gap as a
calibrated leakage measurement, is a concrete direction for keeping reflective optimization auditable
in deployment.

\section{Conclusion}
\label{sec:conclusion}

Reflective prompt optimization delivered a statistically significant, generalizing improvement to a
deep-research agent's citation reward. The held-out gain was 0.079 on 500 topic-disjoint tasks with
no weight updates, it replicated across 3 seeds at a mean of 0.107, and it spread across the
citation-quality levers rather than concentrating in one. Factual support, the hardest lever,
carried the largest share. The methodological contribution is the protocol rather than the
optimizer. A topic-disjoint held-out test, a controlled topic-shared ablation, multi-seed
replication, and a verifier ablation together let us ask whether the large validation-minus-held-out
gap is memorization, and the answer is that it is not. The optimizer does write a literal memorized
answer key, nearly half of the best prompt by word count, but that content earns no reward even on
the exact training questions it encodes. The validation over-statement is therefore small-sample
selection under a noisy judge, not memorized reward. Varying the verifier shows what does drive the
gain. Decomposed per-dimension feedback is the causal ingredient, since a scalar reward drives
held-out transfer negative while finer or repeated scoring does not raise it. The recipe the search
discovered is a portable prompting pattern for citation-faithful research agents, namely an
answer-first structure, one claim per citation, verified page content, and corpus-only URLs. A
hand-written version of this recipe recovers most of the held-out gain, which confirms the
improvement is a transferable rule set rather than a property of the search itself. It
complements rather than replaces reinforcement learning, and it generalizes only when the reward is
decomposed and the evaluation is held-out and multi-seed.

\bibliography{references}

@article{gepa2025,
  title        = {{GEPA}: Reflective Prompt Evolution Can Outperform Reinforcement Learning},
  author       = {Agrawal, Lakshya A. and Tan, Shangyin and Soylu, Dilara and Ziems, Noah and Khare, Rishi and Opsahl-Ong, Krista and Singhvi, Arnav and Shandilya, Herumb and Ryan, Michael J. and Jiang, Meng and Potts, Christopher and Sen, Koushik and Dimakis, Alexandros G. and Stoica, Ion and Klein, Dan and Zaharia, Matei and Khattab, Omar},
  journal      = {arXiv preprint arXiv:2507.19457},
  year         = {2025},
  note         = {ICLR 2026 oral. Code: https://github.com/gepa-ai/gepa},
  url          = {https://arxiv.org/abs/2507.19457}
}

@inproceedings{skalse2022,
  title        = {Defining and Characterizing Reward Hacking},
  author       = {Skalse, Joar and Howe, Nikolaus H. R. and Krasheninnikov, Dmitrii and Krueger, David},
  booktitle    = {Advances in Neural Information Processing Systems (NeurIPS)},
  year         = {2022},
  url          = {https://arxiv.org/abs/2209.13085}
}

@article{amodei2016,
  title        = {Concrete Problems in {AI} Safety},
  author       = {Amodei, Dario and Olah, Chris and Steinhardt, Jacob and Christiano, Paul and Schulman, John and Man{\'e}, Dan},
  journal      = {arXiv preprint arXiv:1606.06565},
  year         = {2016},
  url          = {https://arxiv.org/abs/1606.06565}
}

@inproceedings{gao2023overopt,
  title        = {Scaling Laws for Reward Model Overoptimization},
  author       = {Gao, Leo and Schulman, John and Hilton, Jacob},
  booktitle    = {Proceedings of the 40th International Conference on Machine Learning (ICML)},
  year         = {2023},
  url          = {https://arxiv.org/abs/2210.10760}
}

@article{mahmoud2025,
  title        = {Reward Hacking in Rubric-Based Reinforcement Learning},
  author       = {Mahmoud, Anas and Rezaei, MohammadHossein and Wang, Zihao and Gunjal, Anisha and Liu, Bing and He, Yunzhong},
  journal      = {arXiv preprint arXiv:2605.12474},
  year         = {2026},
  eprint       = {2605.12474},
  archivePrefix = {arXiv}
}

@article{pow3r,
  title        = {Not Every Rubric Teaches Equally: Policy-Aware Rubric Rewards for {RLVR}},
  author       = {Tyagi, Utkarsh and Guo, Xingang and Rezaei, MohammadHossein and George, Daniel and Mahmoud, Anas and Lee, Jackson and Liu, Bing and He, Yunzhong},
  journal      = {arXiv preprint arXiv:2605.20164},
  year         = {2026},
  eprint       = {2605.20164},
  archivePrefix = {arXiv}
}

@article{rgsd,
  title        = {Rubric-Guided Self-Distillation: Post-Training Without Rubric Verifiers},
  author       = {Rezaei, MohammadHossein and Mahmoud, Anas and Wang, Zihao and Tyagi, Utkarsh and Gosai, Advait and Dumitru, Razvan-Gabriel and Sabharwal, Aakash and Liu, Bing and He, Yunzhong},
  journal      = {arXiv preprint arXiv:2606.12507},
  year         = {2026},
  eprint       = {2606.12507},
  archivePrefix = {arXiv}
}

@article{yang2026skillopt,
  title        = {{SkillOpt}: Executive Strategy for Self-Evolving Agent Skills},
  author       = {Yang, Yifan and Gong, Ziyang and Huang, Weiquan and Yang, Qihao and Zhou, Ziwei and Huang, Zisu and Li, Yan and Gao, Xuemei and Dai, Qi and Liu, Bei and Qiu, Kai and Yang, Yuqing and Chen, Dongdong and Yang, Xue and Luo, Chong},
  journal      = {arXiv preprint arXiv:2605.23904},
  year         = {2026},
  eprint       = {2605.23904},
  archivePrefix = {arXiv}
}

@article{grpo2024,
  title        = {{DeepSeekMath}: Pushing the Limits of Mathematical Reasoning in Open Language Models},
  author       = {Shao, Zhihong and Wang, Peiyi and Zhu, Qihao and Xu, Runxin and Song, Junxiao and Bi, Xiao and Zhang, Haowei and Zhang, Mingchuan and Li, Y. K. and Wu, Y. and Guo, Daya},
  journal      = {arXiv preprint arXiv:2402.03300},
  year         = {2024},
  url          = {https://arxiv.org/abs/2402.03300}
}

@inproceedings{miprov2,
  title        = {Optimizing Instructions and Demonstrations for Multi-Stage Language Model Programs},
  author       = {Opsahl-Ong, Krista and Ryan, Michael J. and Purtell, Josh and Broman, David and Potts, Christopher and Zaharia, Matei and Khattab, Omar},
  booktitle    = {Proceedings of the 2024 Conference on Empirical Methods in Natural Language Processing (EMNLP)},
  year         = {2024},
  url          = {https://dspy.ai}
}

@misc{decagon2026,
  title        = {Optimizing {GEPA} for Production: A Test-Driven Approach to Prompt Engineering},
  author       = {{Decagon}},
  year         = {2026},
  howpublished = {\url{https://decagon.ai/blog/optimizing-gepa-for-production}}
}

@misc{combee2026,
  title        = {Scaling {GEPA} with Combee: Parallel Prompt Learning for Self-Improving Agents},
  author       = {{GEPA Team}},
  year         = {2026},
  howpublished = {\url{https://gepa-ai.github.io/gepa/blog/2026/04/09/gepa-at-scale-with-combee/}}
}

@article{onweller2026,
  title        = {Cited but Not Verified: Parsing and Evaluating Source Attribution in {LLM} Deep Research Agents},
  author       = {Onweller, Hailey and Lumer, Elias and Huber, Austin and Ramchandani, Pia and Subbiah, Vamse Kumar and Feld, Corey},
  journal      = {arXiv preprint arXiv:2605.06635},
  year         = {2026},
  url          = {https://arxiv.org/abs/2605.06635}
}

@article{sen2026,
  title        = {Is Grep All You Need? How Agent Harnesses Reshape Agentic Search},
  author       = {Sen, Sahil and Kasturi, Akhil and Lumer, Elias and Gulati, Anmol and Subbiah, Vamse Kumar},
  journal      = {arXiv preprint arXiv:2605.15184},
  year         = {2026},
  url          = {https://arxiv.org/abs/2605.15184}
}

@article{chronos2026,
  title        = {Chronos: Temporal-Aware Conversational Agents with Structured Event Retrieval for Long-Term Memory},
  author       = {Sen, Sahil and Lumer, Elias and Gulati, Anmol and Subbiah, Vamse Kumar},
  journal      = {arXiv preprint arXiv:2603.16862},
  year         = {2026},
  url          = {https://arxiv.org/abs/2603.16862}
}

@misc{gepacode,
  title        = {{GEPA}: Optimize Prompts, Code, and More with {AI}-Powered Reflective Text Evolution},
  author       = {{GEPA-AI}},
  year         = {2025},
  howpublished = {\url{https://github.com/gepa-ai/gepa}},
  note         = {Software repository}
}

@inproceedings{zhou2022ape,
  title        = {Large Language Models Are Human-Level Prompt Engineers},
  author       = {Zhou, Yongchao and Muresanu, Andrei Ioan and Han, Ziwen and Paster, Keiran and Pitis, Silviu and Chan, Harris and Ba, Jimmy},
  booktitle    = {The Eleventh International Conference on Learning Representations (ICLR)},
  year         = {2023},
  url          = {https://arxiv.org/abs/2211.01910}
}

@article{yang2024opro,
  title        = {Large Language Models as Optimizers},
  author       = {Yang, Chengrun and Wang, Xuezhi and Lu, Yifeng and Liu, Hanxiao and Le, Quoc V. and Zhou, Denny and Chen, Xinyun},
  journal      = {arXiv preprint arXiv:2309.03409},
  year         = {2024},
  url          = {https://arxiv.org/abs/2309.03409}
}

@inproceedings{fernando2024promptbreeder,
  title        = {Promptbreeder: Self-Referential Self-Improvement via Prompt Evolution},
  author       = {Fernando, Chrisantha and Banarse, Dylan and Michalewski, Henryk and Osindero, Simon and Rockt{\"a}schel, Tim},
  booktitle    = {Proceedings of the 41st International Conference on Machine Learning (ICML)},
  year         = {2024},
  url          = {https://arxiv.org/abs/2309.16797}
}

@inproceedings{guo2024evoprompt,
  title        = {Connecting Large Language Models with Evolutionary Algorithms Yields Powerful Prompt Optimizers},
  author       = {Guo, Qingyan and Wang, Rui and Guo, Junliang and Li, Bei and Song, Kaitao and Tan, Xu and Liu, Guoqing and Bian, Jiang and Yang, Yujiu},
  booktitle    = {The Twelfth International Conference on Learning Representations (ICLR)},
  year         = {2024},
  url          = {https://openreview.net/forum?id=ZG3RaNIsO8}
}

@misc{agarwal2024promptwizard,
  title        = {PromptWizard: Task-Aware Prompt Optimization Framework},
  author       = {Agarwal, Eshaan and Singh, Joykirat and Dani, Vivek and Magazine, Raghav and Ganu, Tanuja and Nambi, Akshay},
  year         = {2024},
  howpublished = {\url{https://arxiv.org/abs/2405.18369}}
}

@inproceedings{wan2024,
  title        = {Teach Better or Show Smarter? On Instructions and Exemplars in Automatic Prompt Optimization},
  author       = {Wan, Xingchen and Sun, Ruoxi and Nakhost, Hootan and Arik, Sercan},
  booktitle    = {Advances in Neural Information Processing Systems (NeurIPS)},
  volume       = {37},
  year         = {2024}
}

@inproceedings{khattab2024dspy,
  title        = {{DSPy}: Compiling Declarative Language Model Calls into Self-Improving Pipelines},
  author       = {Khattab, Omar and Singhvi, Arnav and Maheshwari, Paridhi and Zhang, Zhiyuan and Santhanam, Keshav and Vardhamanan, Sri and Haq, Saiful and Sharma, Ashutosh and Joshi, Thomas T. and Moazam, Hanna and Miller, Heather and Zaharia, Matei and Potts, Christopher},
  booktitle    = {The Twelfth International Conference on Learning Representations (ICLR)},
  year         = {2024},
  url          = {https://openreview.net/forum?id=sY5N0zY5Od}
}

@article{khattab2022dsp,
  title        = {Demonstrate-Search-Predict: Composing Retrieval and Language Models for Knowledge-Intensive {NLP}},
  author       = {Khattab, Omar and Santhanam, Keshav and Li, Xiang Lisa and Hall, David and Liang, Percy and Potts, Christopher and Zaharia, Matei},
  journal      = {arXiv preprint arXiv:2212.14024},
  year         = {2022},
  url          = {https://arxiv.org/abs/2212.14024}
}

@misc{soylu2024,
  title        = {Fine-Tuning and Prompt Optimization: Two Great Steps That Work Better Together},
  author       = {Soylu, Dilara and Potts, Christopher and Khattab, Omar},
  year         = {2024},
  howpublished = {\url{https://arxiv.org/abs/2407.10930}}
}

@misc{wu2025optimas,
  title        = {Optimas: Optimizing Compound {AI} Systems with Globally Aligned Local Rewards},
  author       = {Wu, Shirley and Sarthi, Parth and Zhao, Shiyu and Lee, Aaron and Shandilya, Herumb and Grobelnik, Adrian Mladenic and Choudhary, Nurendra and Huang, Eddie and Subbian, Karthik and Zhang, Linjun and Yang, Diyi and Zou, James and Leskovec, Jure},
  year         = {2025},
  howpublished = {\url{https://arxiv.org/abs/2507.03041}}
}

@misc{tan2025langprobe,
  title        = {{LangProBe}: A Language Programs Benchmark},
  author       = {Tan, Shangyin and Agrawal, Lakshya A. and Singhvi, Arnav and Lai, Liheng and Ryan, Michael J. and Klein, Dan and Khattab, Omar and Sen, Koushik and Zaharia, Matei},
  year         = {2025},
  howpublished = {\url{https://arxiv.org/abs/2502.20315}}
}

@article{ziems2025mmgrpo,
  title        = {Multi-Module {GRPO}: Composing Policy Gradients and Prompt Optimization for Language Model Programs},
  author       = {Ziems, Noah and Soylu, Dilara and Agrawal, Lakshya A. and Miller, Isaac and Lai, Liheng and Qian, Chen and Song, Kaiqiang and Jiang, Meng and Klein, Dan and Zaharia, Matei and D'Oosterlinck, Karel and Potts, Christopher and Khattab, Omar},
  journal      = {arXiv preprint arXiv:2508.04660},
  year         = {2025},
  url          = {https://arxiv.org/abs/2508.04660}
}

@article{yuksekgonul2025textgrad,
  title        = {Optimizing Generative {AI} by Backpropagating Language Model Feedback},
  author       = {Yuksekgonul, Mert and Bianchi, Federico and Boen, Joseph and Liu, Sheng and Lu, Pan and Huang, Zhi and Guestrin, Carlos and Zou, James},
  journal      = {Nature},
  volume       = {639},
  pages        = {609--616},
  year         = {2025}
}

@misc{xu2025langfeedback,
  title        = {Formalizing Learning from Language Feedback with Provable Guarantees},
  author       = {Xu, Wanqiao and Nie, Allen and Zheng, Ruijie and Modi, Aditya and Swaminathan, Adith and Cheng, Ching-An},
  year         = {2025},
  howpublished = {\url{https://arxiv.org/abs/2506.10341}}
}

@misc{suzgun2025cheatsheet,
  title        = {Dynamic Cheatsheet: Test-Time Learning with Adaptive Memory},
  author       = {Suzgun, Mirac and Yuksekgonul, Mert and Bianchi, Federico and Jurafsky, Dan and Zou, James},
  year         = {2025},
  howpublished = {\url{https://arxiv.org/abs/2504.07952}}
}

@misc{wang2024awm,
  title        = {Agent Workflow Memory},
  author       = {Wang, Zora Zhiruo and Mao, Jiayuan and Fried, Daniel and Neubig, Graham},
  year         = {2024},
  howpublished = {\url{https://arxiv.org/abs/2409.07429}}
}

@misc{yao2023react,
  title        = {{ReAct}: Synergizing Reasoning and Acting in Language Models},
  author       = {Yao, Shunyu and Zhao, Jeffrey and Yu, Dian and Du, Nan and Shafran, Izhak and Narasimhan, Karthik and Cao, Yuan},
  year         = {2023},
  note         = {ICLR 2023},
  howpublished = {\url{https://arxiv.org/abs/2210.03629}}
}

@misc{wei2023cot,
  title        = {Chain-of-Thought Prompting Elicits Reasoning in Large Language Models},
  author       = {Wei, Jason and Wang, Xuezhi and Schuurmans, Dale and Bosma, Maarten and Ichter, Brian and Xia, Fei and Chi, Ed and Le, Quoc and Zhou, Denny},
  year         = {2023},
  note         = {NeurIPS 2022},
  howpublished = {\url{https://arxiv.org/abs/2201.11903}}
}

@misc{jin2025searchr1,
  title        = {Search-R1: Training {LLMs} to Reason and Leverage Search Engines with Reinforcement Learning},
  author       = {Jin, Bowen and Zeng, Hansi and Yue, Zhenrui and Yoon, Jinsung and Arik, Sercan and Wang, Dong and Zamani, Hamed and Han, Jiawei},
  year         = {2025},
  howpublished = {\url{https://arxiv.org/abs/2503.09516}}
}

@misc{chen2025research,
  title        = {{ReSearch}: Learning to Reason with Search for {LLMs} via Reinforcement Learning},
  author       = {Chen, Mingyang and Sun, Linzhuang and Li, Tianpeng and Sun, Haoze and Zhou, Yijie and Zhu, Chenzheng and Wang, Haofen and Pan, Jeff Z. and Zhang, Wen and Chen, Huajun and Yang, Fan and Zhou, Zenan and Chen, Weipeng},
  year         = {2025},
  howpublished = {\url{https://arxiv.org/abs/2503.19470}}
}

@misc{song2025r1searcher,
  title        = {R1-Searcher: Incentivizing the Search Capability in {LLMs} via Reinforcement Learning},
  author       = {Song, Huatong and Jiang, Jinhao and Min, Yingqian and Chen, Jie and Chen, Zhipeng and Zhao, Wayne Xin and Fang, Lei and Wen, Ji-Rong},
  year         = {2025},
  howpublished = {\url{https://arxiv.org/abs/2503.05592}}
}

@misc{java2025frugalrag,
  title        = {{FrugalRAG}: Less is More in {RL} Finetuning for Multi-Hop Question Answering},
  author       = {Java, Abhinav and Koundinyan, Srivathsan and Natarajan, Nagarajan and Sharma, Amit},
  year         = {2025},
  howpublished = {\url{https://arxiv.org/abs/2507.07634}}
}

@misc{wu2025searchwisely,
  title        = {Search Wisely: Mitigating Sub-Optimal Agentic Searches by Reducing Uncertainty},
  author       = {Wu, Peilin and Zhang, Mian and Zhang, Xinlu and Du, Xinya and Chen, Zhiyu Zoey},
  year         = {2025},
  howpublished = {\url{https://arxiv.org/abs/2505.17281}}
}

@misc{pyatkin2025vif,
  title        = {Generalizing Verifiable Instruction Following},
  author       = {Pyatkin, Valentina and Malik, Saumya and Graf, Victoria and Ivison, Hamish and Huang, Shengyi and Dasigi, Pradeep and Lambert, Nathan and Hajishirzi, Hannaneh},
  year         = {2025},
  howpublished = {\url{https://arxiv.org/abs/2507.02833}}
}

@misc{peng2025verif,
  title        = {{VerIF}: Verification Engineering for Reinforcement Learning in Instruction Following},
  author       = {Peng, Hao and Qi, Yunjia and Wang, Xiaozhi and Xu, Bin and Hou, Lei and Li, Juanzi},
  year         = {2025},
  howpublished = {\url{https://arxiv.org/abs/2506.09942}}
}

@misc{si2025faithfulness,
  title        = {Teaching Large Language Models to Maintain Contextual Faithfulness via Synthetic Tasks and Reinforcement Learning},
  author       = {Si, Shuzheng and Zhao, Haozhe and Gao, Cheng and Bai, Yuzhuo and Wang, Zhitong and Gao, Bofei and Luo, Kangyang and Li, Wenhao and Huang, Yufei and Chen, Gang and Qi, Fanchao and Zhang, Minjia and Chang, Baobao and Sun, Maosong},
  year         = {2025},
  howpublished = {\url{https://arxiv.org/abs/2505.16483}}
}

@misc{sun2025hallucination,
  title        = {Detection and Mitigation of Hallucination in Large Reasoning Models: A Mechanistic Perspective},
  author       = {Sun, Zhongxiang and Wang, Qipeng and Wang, Haoyu and Zhang, Xiao and Xu, Jun},
  year         = {2025},
  howpublished = {\url{https://arxiv.org/abs/2505.12886}}
}

@inproceedings{yang2018hotpotqa,
  title        = {{HotpotQA}: A Dataset for Diverse, Explainable Multi-Hop Question Answering},
  author       = {Yang, Zhilin and Qi, Peng and Zhang, Saizheng and Bengio, Yoshua and Cohen, William W. and Salakhutdinov, Ruslan and Manning, Christopher D.},
  booktitle    = {Proceedings of the 2018 Conference on Empirical Methods in Natural Language Processing (EMNLP)},
  year         = {2018}
}

@inproceedings{jiang2020hover,
  title        = {{HoVer}: A Dataset for Many-Hop Fact Extraction and Claim Verification},
  author       = {Jiang, Yichen and Bordia, Shikha and Zhong, Zheng and Dognin, Charles and Singh, Maneesh and Bansal, Mohit},
  booktitle    = {Findings of the Association for Computational Linguistics: EMNLP 2020},
  pages        = {3441--3460},
  year         = {2020}
}

@misc{leung2026needfrontiermodelcitation,
  title         = {Do You Need a Frontier Model as a Citation Verifier? Benchmarking Rubric {LLMs} for Deep-Research Source Attribution},
  author        = {Leung, Ethan and Lumer, Elias and Feld, Corey and Huber, Austin and Subbiah, Vamse Kumar and Paul, Kevin},
  year          = {2026},
  eprint        = {2607.08700},
  archivePrefix = {arXiv},
  primaryClass  = {cs.CL},
  url           = {https://arxiv.org/abs/2607.08700}
}

@article{lumer2025multimodalrag,
  title        = {Comparison of Text-Based and Image-Based Retrieval in Multimodal Retrieval Augmented Generation Large Language Model Systems},
  author       = {Lumer, Elias and Cardenas, Alex and Melich, Matt and Mason, Myles and Dieter, Sara and Subbiah, Vamse Kumar and Basavaraju, Pradeep Honaganahalli and Hernandez, Roberto},
  journal      = {arXiv preprint arXiv:2511.16654},
  year         = {2025},
  url          = {https://arxiv.org/abs/2511.16654}
}

@article{lumer2026rah,
  title        = {Recursive Agent Harnesses},
  author       = {Lumer, Elias and Sen, Sahil and Paul, Kevin and Subbiah, Vamse Kumar},
  journal      = {arXiv preprint arXiv:2606.13643},
  year         = {2026},
  url          = {https://arxiv.org/abs/2606.13643}
}

@article{reward_survey,
  title        = {Reward Hacking in the Era of Large Models: Mechanisms, Emergent Misalignment, Challenges},
  author       = {Wang, Xiaohua and Tian, Muzhao and Zeng, Yuqi and Huang, Zisu and Yuan, Jiakang and Chen, Bowen and Xu, Jingwen and Zhou, Mingbo and Liu, Wenhao and Wu, Muling and Guo, Zhengkang and Qian, Qi and Wang, Yifei and Zhang, Feiran and Yin, Ruicheng and Dou, Shihan and Lv, Changze and Chen, Tao and Song, Kaitao and Tan, Xu and Gui, Tao and Zheng, Xiaoqing and Huang, Xuanjing},
  journal      = {arXiv preprint arXiv:2604.13602},
  year         = {2026},
  url          = {https://arxiv.org/abs/2604.13602}
}

@inproceedings{coste2024ensembles,
  title        = {Reward Model Ensembles Help Mitigate Overoptimization},
  author       = {Coste, Thomas and Anwar, Usman and Kirk, Robert and Krueger, David},
  booktitle    = {The Twelfth International Conference on Learning Representations (ICLR)},
  year         = {2024},
  url          = {https://arxiv.org/abs/2310.02743}
}

@article{manheim2018goodhart,
  title        = {Categorizing Variants of {Goodhart's} Law},
  author       = {Manheim, David and Garrabrant, Scott},
  journal      = {arXiv preprint arXiv:1803.04585},
  year         = {2018},
  url          = {https://arxiv.org/abs/1803.04585}
}

@inproceedings{zheng2023judge,
  title        = {Judging {LLM}-as-a-Judge with {MT-Bench} and Chatbot Arena},
  author       = {Zheng, Lianmin and Chiang, Wei-Lin and Sheng, Ying and Zhuang, Siyuan and Wu, Zhanghao and Zhuang, Yonghao and Lin, Zi and Li, Zhuohan and Li, Dacheng and Xing, Eric P. and Zhang, Hao and Gonzalez, Joseph E. and Stoica, Ion},
  booktitle    = {Advances in Neural Information Processing Systems (NeurIPS), Datasets and Benchmarks Track},
  year         = {2023},
  url          = {https://arxiv.org/abs/2306.05685}
}

@inproceedings{panickssery2024selfpref,
  title        = {{LLM} Evaluators Recognize and Favor Their Own Generations},
  author       = {Panickssery, Arjun and Bowman, Samuel R. and Feng, Shi},
  booktitle    = {Advances in Neural Information Processing Systems (NeurIPS)},
  year         = {2024},
  url          = {https://arxiv.org/abs/2404.13076}
}

@article{efron1979bootstrap,
  title        = {Bootstrap Methods: Another Look at the Jackknife},
  author       = {Efron, Bradley},
  journal      = {The Annals of Statistics},
  volume       = {7},
  number       = {1},
  pages        = {1--26},
  year         = {1979}
}

@article{wilcoxon1945,
  title        = {Individual Comparisons by Ranking Methods},
  author       = {Wilcoxon, Frank},
  journal      = {Biometrics Bulletin},
  volume       = {1},
  number       = {6},
  pages        = {80--83},
  year         = {1945}
}

@article{cohen1960kappa,
  title        = {A Coefficient of Agreement for Nominal Scales},
  author       = {Cohen, Jacob},
  journal      = {Educational and Psychological Measurement},
  volume       = {20},
  number       = {1},
  pages        = {37--46},
  year         = {1960}
}

@article{fu2026textreg,
  title        = {{TextReg}: Mitigating Prompt Distributional Overfitting via Regularized Text-Space Optimization},
  author       = {Fu, Lucheng and Yu, Ye and Wang, Yiyang and Jin, Yiqiao and Jin, Haibo and Prakash, B. Aditya and Wang, Haohan},
  journal      = {arXiv preprint arXiv:2605.21318},
  year         = {2026},
  url          = {https://arxiv.org/abs/2605.21318}
}

@article{bai2026maspromptbench,
  title        = {{MAS-PromptBench}: When Does Prompt Optimization Improve Multi-Agent {LLM} Systems?},
  author       = {Bai, Juyang and Shi, Laixi},
  journal      = {arXiv preprint arXiv:2606.23664},
  year         = {2026},
  url          = {https://arxiv.org/abs/2606.23664}
}

@article{xu2026traindr,
  title        = {How to Train Your Deep Research Agent? Prompt, Reward, and Policy Optimization in {Search-R1}},
  author       = {Xu, Yinuo and Lu, Shuo and Cheng, Jianjie and Wang, Meng and Xie, Qianlong and Wang, Xingxing and He, Ran and Liang, Jian},
  journal      = {arXiv preprint arXiv:2602.19526},
  year         = {2026},
  url          = {https://arxiv.org/abs/2602.19526}
}

@article{shi2025deepresearch,
  title        = {Deep Research: A Systematic Survey},
  author       = {Shi, Zhengliang and Chen, Yiqun and Li, Haitao and Sun, Weiwei and Ni, Shiyu and Lyu, Yougang and Fan, Run-Ze and Jin, Bowen and Weng, Yixuan and Zhu, Minjun and Xie, Qiujie and Guo, Xinyu and Yang, Qu and Wu, Jiayi and Zhao, Jujia and Tang, Xiaqiang and Ma, Xinbei and Wang, Cunxiang and Mao, Jiaxin and Ai, Qingyao and Huang, Jen-Tse and Wang, Wenxuan and Zhang, Yue and Yang, Yiming and Tu, Zhaopeng and Ren, Zhaochun},
  journal      = {arXiv preprint arXiv:2512.02038},
  year         = {2025},
  url          = {https://arxiv.org/abs/2512.02038}
}

@article{wang2026cherrl,
  title        = {Reproducing, Analyzing, and Detecting Reward Hacking in Rubric-Based Reinforcement Learning},
  author       = {Wang, Xuekang and Hao, Zhuoyuan and Hou, Shuo and Peng, Hao and Li, Juanzi and Wang, Xiaozhi},
  journal      = {arXiv preprint arXiv:2606.04923},
  year         = {2026},
  url          = {https://arxiv.org/abs/2606.04923}
}

@article{yang2026rubricdropout,
  title        = {Rubric Dropout: A Simple Way to Mitigate Reward Hacking in Rubric-as-Reward {RL}},
  author       = {Yang, Minglai and Guo, Xinyu and Tyagi, Utkarsh and Zhang, Mian and Dumitru, Razvan and Hou, Sunjie and He, Yunzhong and Zhang, Daniel Yue and Liu, Ying},
  journal      = {arXiv preprint arXiv:2608.11669},
  year         = {2026},
  url          = {https://arxiv.org/abs/2608.11669}
}

@article{pombal2026selfpref,
  title        = {Self-Preference Bias in Rubric-Based Evaluation of Large Language Models},
  author       = {Pombal, Jos{\'e} and Rei, Ricardo and Martins, Andr{\'e} F. T.},
  journal      = {arXiv preprint arXiv:2604.06996},
  year         = {2026},
  url          = {https://arxiv.org/abs/2604.06996}
}

@article{yang2026hackprobe,
  title        = {Harness-Agnostic Detection and Immunization of Reward Hacking in Self-Evolving Language Models},
  author       = {Yang, Rongxin and Liu, Yang and Luo, Shang and Jia, Haoxuan and Zhang, Chongyang and Zheng, Hao and Yang, Yingguang and Huang, Yulin and Zhang, Jianshen and Qi, Yongzhi and Xu, Kefu and Ran, Congjing and Chong, Bin},
  journal      = {arXiv preprint arXiv:2609.04665},
  year         = {2026},
  url          = {https://arxiv.org/abs/2609.04665}
}

\appendix

\section{Reproducibility}
\label{app:repro}

The analysis runs as a pipeline over the released artifact. The optimization step runs the agent
with the seed prompts on the train and validation splits and searches with GEPA under the
configuration in Section~\ref{sec:gepa}. The analysis step re-scores every candidate on
validation and re-scores the seed and best prompts on the 200-task held-out subset with
per-citation reasons enabled. The reporting step derives the prompt lineage and diffs, the paired
held-out statistics, the figures, the tables, and the error analysis. The commands below are
illustrative and use the released artifact layout.

\begin{quote}\small\ttfamily
\# 1. Optimize (the run analysed here)\\
optimize --dataset dataset.jsonl --index corpus.db --out run/ \textbackslash\\
\phantom{xx}--components both --train-size 24 --val-size 24 --test-size 500 \textbackslash\\
\phantom{xx}--max-metric-calls 240 --reflection-effort high\\[4pt]
\# 2. Re-analysis (trajectory, decomposition, error material)\\
analyze --run-dir run/ --dataset dataset.jsonl --index corpus.db \textbackslash\\
\phantom{xx}--out data/ --test-decomp-size 200\\[4pt]
\# 3. Lineage and diffs, paired stats, figures, tables, error analysis\\
report --run-dir run/ --data-dir data/ --out .
\end{quote}

\section{Artifacts}
\label{app:artifacts}

We release the seed and all candidate prompts, their pairwise lineage diffs, the machine-readable
lineage in \texttt{data/prompt\_lineage.json}, the per-lever analysis data, the figures, and the
tables, with the prompt-evolution and error-analysis write-ups in \texttt{analysis/} (paths are
given in the \texttt{README}).

\section{Reward weights, answer gate, and judge prompts}
\label{app:reward}

This appendix specifies the gated composite reward of Equation~\eqref{eq:reward} exactly as used in
the run analysed here and in every re-evaluation.

\paragraph{Composite weights and aggregation.} The four weights are
$w_{\text{link}} = 0.25$, $w_{\text{rel}} = 0.30$, $w_{\text{fact}} = 0.30$, and
$w_{\text{num}} = 0.15$, which sum to 1.00. Each of \texttt{link\_works},
\texttt{relevant\_content}, and \texttt{fact\_check} is the unweighted mean over every parsed
(attribution, citation) pair in the report, and a report with no parseable citation pair scores 0
on all three. Two levers are deterministic rather than model-scored. \texttt{link\_works} is 1 when
the cited URL resolves to content in the frozen corpus, and \texttt{citation\_numbering} rewards
inline \texttt{[n](url)} citations numbered $1..N$ contiguously with a matching \texttt{Sources}
list, penalizing gaps, duplicates, and mismatches. All judges see frozen-corpus content only, with
no live web fetch at reward time.

\paragraph{Answer gate.} The gate is a stateless, cheapest-first cascade returning
$\{0,1\}$. Tier 1 is a deterministic number-canonicalized substring match of the gold answer in the
report (thousands separators stripped, unit forms such as ``41 billion'', ``41bn'', and ``41B''
unified, quotes and dashes normalized, whitespace collapsed, lowercased); a multi-part gold answer
joined by \texttt{;} or \texttt{|} passes only if all parts are present. Tier 2 is a deterministic
coverage router on the fraction of the gold answer's distinctive tokens present in the report, with
a pass threshold of 1.0 and a fail threshold of 0.5. Coverage at or above 1.0 passes, coverage below
0.5 fails with no language-model call, and the near-miss band in between escalates to Tier 3. Tier 3
fires only in that near-miss band and uses OpenAI GPT-5-mini (\texttt{gpt-5-mini-2025-08-07}, max
256 tokens, report truncated to 6000 characters), reasoning before emitting a final
\texttt{VERDICT: yes} or \texttt{VERDICT: no} line. Its verbatim prompt is:

\begin{quote}\sloppy\small\ttfamily
You judge whether a research REPORT answers the QUESTION with the same fact as the GOLD ANSWER.
Paraphrase, restatement, and rounding-free equivalents count as the same fact; a different or absent
fact does not.\\
First reason in 1-3 short sentences about whether the report expresses the gold fact. Then, on a
FINAL separate line, output exactly 'VERDICT: yes' or 'VERDICT: no'.\\[2pt]
QUESTION: \{question\}\\
GOLD ANSWER: \{truth\}\\
REPORT: \{report[:6000]\}
\end{quote}

\paragraph{Citation judges.} \sloppy The source-relevance and factual-support levers are binary
($0/1$) language-model judges that return a structured \texttt{\{score, reason\}} object and see the
cited URL's corpus content truncated to 5000 characters. Source relevance uses OpenAI GPT-5-mini
(\texttt{gpt-5-mini-2025-08-07}) and factual support uses OpenAI GPT-5.4-mini
(\texttt{gpt-5.4-mini-2026-03-17}), following the per-dimension picks of
\citet{leung2026needfrontiermodelcitation}. The verbatim system prompts are:

\begin{quote}\sloppy\small\ttfamily
\textbf{Source relevance (OpenAI GPT-5-mini).}
You are an expert evaluator assessing content relevance. Your task is to determine if the content at
a URL is relevant to a given attribution text. Consider whether the URL content discusses the same
topic as the attribution, provides supporting information for its claims, and has a clear connection
to it. This is a binary evaluation: score = 1 if the content IS relevant, score = 0 if it is NOT.
\\[4pt]
\textbf{Factual support (OpenAI GPT-5.4-mini).}
You are an expert fact-checker evaluating factual accuracy. Your task is to determine if the factual
claims in an attribution text are supported by the content at a URL. Consider whether the specific
facts, numbers, dates, and claims in the attribution are present in the URL content, whether they
match or are consistent, and whether there are any contradictions. This is a binary evaluation:
score = 1 if the facts ARE supported, score = 0 if they are NOT (contradicted, missing, or
unverifiable).
\end{quote}

Each judge is called with the attribution text, the cited URL, and the truncated URL content, and is
asked to evaluate relevance or factual support for that pair.

\end{document}